\documentclass[journal]{IEEEtran}

\usepackage{booktabs}
\usepackage{graphicx}
\usepackage{amsmath}
\usepackage{arydshln}
\usepackage[utf8]{inputenc}
\usepackage[french]{nomencl}
\usepackage{subfigure}
\usepackage{amssymb}
\usepackage{xcolor}
\usepackage{nomencl}
\usepackage{cite}
\usepackage{mathrsfs}
\usepackage{cases}
\usepackage{float}
\usepackage[ruled,vlined]{algorithm2e}

\usepackage{multirow}

\newtheorem{remark}{Remark}
\makenomenclature

\newcommand{\tabincell}[2]{\begin{tabular}{@{}#1@{}}#2\end{tabular}}

\ifCLASSINFOpdf

\else

\fi

\begin{document}

%
\title{Cosserat-Rod Based Dynamic Modeling of Soft Slender Robot Interacting with Environment}
%
%
%

\author{Lingxiao~Xun,~Gang~Zheng,~\IEEEmembership{Senior~Member,~IEEE},~Alexandre Kruszewski

\thanks{Lingxiao Xun, Gang Zheng, Alexandre Kruszewski are in Defrost team, Inria, university of Lille, Centrale Lille, CRIStAL - Centre de Recherche en Informatique Signal et Automatique de Lille - UMR 9189, France (e-mail: lingxiao.xun@inria.fr; gang.zheng@inria.fr; alexandre.kruszewski@centralelille.fr).   }
}
%
%

\markboth{}%
{Shell \MakeLowercase{\textit{et al.}}: Bare Demo of IEEEtran.cls for IEEE Journals}
%



\maketitle

\begin{abstract}
Soft slender robots have attracted more and more research attentions in these years due to their continuity and compliance natures. 
However, mechanics modeling for soft robots interacting with environment is still an academic challenge because of the non-linearity of deformation and the non-smooth property of the contacts.
In this work, starting from a piece-wise local strain field assumption, we propose a nonlinear dynamic model for soft robot via Cosserat rod theory using Newtonian mechanics which handles the frictional contact with environment and transfer them into the nonlinear complementary constraint (NCP) formulation. Moreover, we smooth both the contact and friction constraints in order to convert the inequality equations of  NCP to the smooth equality equations. The proposed model allows us to compute the dynamic deformation and frictional contact force under common optimization framework in real time when the soft slender robot interacts with other rigid or soft bodies. In the end, the corresponding experiments are carried out which valid our proposed dynamic model.
\end{abstract}


%
\IEEEpeerreviewmaketitle

\section{Introduction}
%
%
%
Soft robots possess the ability to adapt to complex external environments by utilizing material deformations and structural changes. This unique characteristic has led to an increasing interest among scholars in recent years regarding the modeling and control of soft robots \cite{laschi2016soft}\cite{trivedi2008soft}\cite{rus2015design}. Unlike traditional rigid structures, soft structures exhibit infinite degrees of freedom, making kinematics and dynamics analysis challenging. Additionally, in contrast to traditional continuum computational mechanics, which primarily emphasizes high simulation accuracy, modeling soft robots requires a balance between real-time performance and a certain level of accuracy to enable real-time extraction of model information during control.

The key challenge in soft robot modeling lies in dealing with its large deformation and computational complexity, which arise due to its high dimensionality. Thus, the pursuit of efficient and reliable models has become a crucial research direction in soft robotics. It is important to note that in most robotic applications, real-time interaction with the external world is essential, such as robot locomotion, manipulation tasks, and medical operations within the human body such as endoscopic surgery, aneurysm surgery, and cochlear implantation \cite{dogangil2010review}\cite{calisti2017fundamentals}\cite{cianchetti2018biomedical}. While in these applications, contact deformation and contact forces between objects are indispensable and important factors in the design and control of soft robots, thus the contact modeling plays a pivotal role in the overall analysis. However, while these aspects have been extensively studied in the field of rigid body robotics, they have not yet received sufficient attention in the realm of soft robotics. In the following review, we will discuss relevant works focusing on the mechanical and contact models of soft robots separately.
\subsection{Soft slender robot modeling}
The mechanical modeling of soft robots falls within the realm of continuum mechanics. When dealing with objects that possess complex geometric structures, the finite element method (FEM) is commonly employed, which has found widespread applications in engineering practice over the past few decades\cite{bieze2018finite}. However, for soft robots with slender geometric structures, they can be treated as single beams or combinations of multi-body beams. As a result, the modeling of many soft and slender robots can be approximated using beam models with lower dimensions. In recent years, beam models have been extensively utilized in the modeling of various soft robots, yielding promising outcomes \cite{olson2020euler}.

Beam/rod theory is indeed a subclass of continuum mechanics that has found wide application in engineering mechanics simulations due to its universal structure and mathematical simplicity. The classical Euler-Bernoulli beam theory has been extensively used in the mechanical modeling of slender robot structures. For instance, in the study by Olson et al. \cite{olson2020euler}, a quasi-static bending model based on the geometrically accurate Euler-Bernoulli formulation was developed, allowing for the prediction of design outcomes for various soft-arm designs. Another commonly employed beam model in soft robot modeling is the Timoshenko beam model, which takes into account the influence of shear strain \cite{lindenroth2016stiffness}\cite{godaba2019payload}.

In addition to the Euler-Bernoulli and Timoshenko beam models, the Kirchhoff rod model is an extension that considers the torsional strain of the beam \cite{boyer2011geometrically}\cite{novelia2018discrete}. When dealing with large deformations of slender soft robots, geometrically nonlinear classical rod theory is often employed. The pseudo-rigid body (PRB) 3R model was initially developed to capture large deformations of flexible beams subjected to tip loading and demonstrated high computational efficiency in analyzing compliant mechanisms \cite{su2009pseudorigid}. Building upon the PRB 3R model, Huang et al. \cite{huang20193d} proposed a three-dimensional (3D) static modeling method for cable-actuated continuum robots, aiming to extend the PRB 3R model to 3D applications and to achieve high model accuracy.

In many continuum models, constant curvature (CC) assumptions are frequently employed to approximate large deformations, simplifying the computational aspects of the mathematical models. This motivates the utilization of generalized elastic models, such as the Cosserat rod theory, in various applications \cite{boyer2017poincare}\cite{black2017parallel}\cite{rucker2011statics}\cite{cao2008nonlinear}\cite{haibin2018modeling}. Cosserat rods, which are geometrically nonlinear generalizations of Timoshenko-Reissner beams, offer the capability to simulate bending, torsion, shear, and tension in soft-body continuum robots. A detailed review on the statics, dynamics, and stability of Cosserat-based slender elastic rod continuum robots is provided in \cite{till2019statics}. Building upon this foundation, new methods for solving Cosserat partial differential equations (PDEs) in continuous space have been proposed in \cite{till2019real} based on Newton-Euler dynamics.

In the context of soft slender robots, the work in \cite{zhang2019modeling} introduces a Cosserat-based piecewise constant strain model where the PDEs are transformed into an approximate weak form expressed as ordinary differential equations (ODEs). Another approach, presented in \cite{boyer2020dynamics}, utilizes a Cosserat discrete solution method based on strain nonlinear parameterization and Lagrangian dynamics. However, this technique can become computationally complex when modeling involves complex deformations like buckling behavior and local strain variations due to contact. To address this challenge, we will employ a piecewise local approximation of the strain field in our work.

The Cosserat model has also been applied in specific engineering scenarios. In \cite{renda2014dynamic}, a Cosserat-based geometrically accurate dynamics model was developed for a soft manipulator actuated by cables. Another study \cite{zhang2019modeling} introduces a method that assembles heterogeneous, active, and passive Cosserat rods to simulate dynamic musculoskeletal structures capable of withstanding all deformation modes. These examples demonstrate the practical applications of the Cosserat model in the field of soft robotics.
\subsection{Frictional contact}
The solution to contact problems in continuum mechanics has been a subject of considerable interest in recent decades, with a wide range of contact models being employed in engineering fields. However, solving the problem of multiple frictional contacts with large deformations remains challenging \cite{stewart2000rigid}. Although contact research for flexible systems has made advancements \cite{wasfy2003computational}, the development of efficient and robust frictional contact algorithms still present open challenges.

The penalty function method is widely used in computational contact mechanics due to its simplicity and directness \cite{munjiza2000penalty}. However, stiffness and stability issues persist despite recent progress \cite{spillmann2007non}. Numerical optimization-based contact solution methods often involve linear complementary programming (LCP) or nonlinear complementary programming (NCP) formulations, which offer higher accuracy at the cost of increased computational complexity and the use of frictional approximations \cite{milenkovic2001optimization}\cite{kaufman2005fast}. LCP can be solved using relaxation methods like projected Gauss-Seidel (PGS) or direct methods such as Dantzig's pivoting algorithm or Lemke's algorithm. In the work of Stewart and Trinkle \cite{stewart1996implicit}, the Coulomb friction cone is linearized, and Lemke's method is used to solve the resulting polygonal pyramid LCP, handling nonlinear forces. Subsequently, \cite{erleben2007velocity} proposed a method based on the Gaussian Seidel iterative inverse solution to LCP, which introduced a solution framework to the field of deformable solids \cite{duriez2005realistic} \cite{otaduy2009implicit}. \cite{harmon2009asynchronous} introduced a contact model for asynchronously handling deformable solids by discretizing the contact barrier potential. For a comprehensive review of numerical methods for linear complementarity problems, we refer to \cite{niebe2015numerical}.

In addition to LCP formulations, another common approach to solve contact problems is through Newton-type methods, which typically utilize the generalized projection operator and augmented Lagrangian to address contact constraints \cite{alart1991mixed}\cite{curnier1988generalized}\cite{dirkse1995path}. Recent works have applied these methods in various applications \cite{bertails2011nonsmooth}\cite{daviet2011hybrid}\cite{kaufman2014adaptive}. The complementarity problem in these works is addressed using non-smooth functions and solved from the perspective of a common Newton's method. This Newton-type optimization ensures quadratic convergence, although the number of iterations is typically higher compared to relaxation methods. For a comprehensive review of non-smooth methods applied to dynamics problems, we refer to \cite{acary2008numerical}.
\subsection{Contributions and outline}
Building on the related work and limitations discussed above, the contributions of our work in this paper are as follows: 

(1). This work aims to establish a general framework for modeling the dynamics of soft slender robots via Cosserat rod theory to handle frictional contact. Precisely, we proposed a novel approach for modeling the dynamics of soft slender robots through piecewise linear interpolation for both strain field and contact field. The proposed model,  profited by relative low dimension and a balance between accuracy, robustness and computational complexity, intents to lay the foundation for real-time simulation and control works for soft slender robot under environment interaction.

(2). We proposed a new formulation for solving the dynamics with frictional contact constraint. In stead of solving LCP or NCP non-smoothly, we reformulate the contact constraint as the equality constraint using the general smooth NCP functions and extend this formulation to handle soft slender robots. This new formulation can be solved by common Newton direct method, which  allows us naturally and efficiently to accommodate nonlinear friction models.  

In summary, this paper is organized as follows: The problem statement is addressed in Section \ref{sec::pro}.
Section \ref{sec::Kinematics} provides introduces the configuration method of soft slender robots based on Lie groups and discusses the representation of contact surfaces. Section \ref{sec::dynamics} and \ref{sec::Discritization} discuss the contact dynamics of soft slender robots in both continuous and discrete forms. In Section \ref{sec::cont_constrain}, we present an overview of contact constraints in soft robotics. Subsequently, Section \ref{Smoothing} introduces the reformulation of contact constraints via equality equations. In Section \ref{sec::implicit}, we propose the formulation of implicit dynamics which can be solved by common Newton method. Section \ref{sec::simu} and \ref{sec::exp} present numerical simulation examples and experimental comparisons based on the proposed model, respectively.
Finally, Section \ref{sec::conc} concludes this article. 
\begin{table}
	\centering
	\caption{Nomenclature and definitions}
	\begin{tabular}{lll}
		\toprule
		\multirow{1}{*}{Symbol} & \multicolumn{1}{l}{\tabincell{l}{Unit }}&\multicolumn{1}{l}{\tabincell{c}{Definition}}\\
		\midrule
		$s$ & ---  & Coordinate of arc length. \\
		$\beta$ & ---  & Coordinate of cross-section. \\
		$t$ & s  & Time. \\
		\multirow{1}{*}{$\boldsymbol{d}(s,\beta)$} & \multicolumn{1}{l}{\tabincell{l}{$\rm m$ }}&\multicolumn{1}{l}{\tabincell{l}{Local distance between the midline of soft \\slender robot and the contact point.}}\\
		$\boldsymbol{R}(s, t)$ & --- & \multicolumn{1}{l}{\tabincell{l}{Rotation matrix with respect to the inertail\\frame.}}\\
		$\boldsymbol{p}(s, t)$ & m & \multicolumn{1}{l}{\tabincell{l}{Position vector with respect to the inertail\\frame.}} \\
		\multirow{1}{*}{$\boldsymbol{g}(s, t)$} & \multicolumn{1}{l}{\tabincell{l}{--- }}&\multicolumn{1}{l}{\tabincell{l}{The configuration tensor of cross-section of \\soft slender robot}}\\
		\multirow{1}{*}{$\boldsymbol{g}_0(s,t)$} & \multicolumn{1}{l}{\tabincell{l}{--- }}&\multicolumn{1}{l}{\tabincell{l}{The transformation matrix from the soft slender\\ robot's base frame to the inertial frame.}}\\
		\multirow{1}{*}{$\boldsymbol{g}_c(s,t)$} & \multicolumn{1}{l}{\tabincell{l}{--- }}&\multicolumn{1}{l}{\tabincell{l}{The configuration tensor of contact frame.}}\\
		\multirow{1}{*}{$\boldsymbol{g}_{bc}(s,t)$} & \multicolumn{1}{l}{\tabincell{l}{--- }}&\multicolumn{1}{l}{\tabincell{l}{the configuration tensor of contact frame with\\respect to body frame.}}\\
		\multirow{1}{*}{$\boldsymbol{g}_d(s,t)$} & \multicolumn{1}{l}{\tabincell{l}{--- }}&\multicolumn{1}{l}{\tabincell{l}{The configuration tensor of slave contact frame.}}\\
		\multirow{1}{*}{$\boldsymbol{g}_{cd}(s,t)$} & \multicolumn{1}{l}{\tabincell{l}{--- }}&\multicolumn{1}{l}{\tabincell{l}{the configuration tensor of slave contact frame\\with respect to master contact frame.}}\\
		\multirow{1}{*}{$\boldsymbol{\kappa}(s, t)$ } & \multicolumn{1}{l}{\tabincell{l}{1/m }}&\multicolumn{1}{l}{\tabincell{l}{Angular strain in the body frame.}}\\
		\multirow{1}{*}{$\boldsymbol{\epsilon}(s, t)$ } & \multicolumn{1}{l}{\tabincell{l}{--- }}&\multicolumn{1}{l}{\tabincell{l}{Linear strain in the body frame.}}\\
		\multirow{1}{*}{$\boldsymbol{\omega}(s, t)$ } & \multicolumn{1}{l}{\tabincell{l}{1/s }}&\multicolumn{1}{l}{\tabincell{l}{Angular velocity expressed in the body frame.}}\\
		\multirow{1}{*}{$\boldsymbol{v}(s, t)$ } & \multicolumn{1}{l}{\tabincell{l}{m/s }}&\multicolumn{1}{l}{\tabincell{l}{Linear velocity expressed in the body frame.}}\\
		\multirow{1}{*}{$\widetilde{(\cdot)}$} & \multicolumn{1}{l}{\tabincell{l}{--- }}&\multicolumn{1}{l}{\tabincell{l}{Mapping from $\mathbb{R}^3$ to $so(3)$, \\e.g. $\scriptsize
				\widetilde{\boldsymbol{a}}=\left[ \begin{matrix}
					0&-a_3&a_2\\
					a_3&0&-a_1\\
					-a_2&a_1&0
				\end{matrix}\right] $.}}\\
		\multirow{1}{*}{$\widehat{(\cdot)}$} & \multicolumn{1}{l}{\tabincell{l}{---}}&\multicolumn{1}{l}{\tabincell{l}{ Mapping from $\mathbb{R}^6$ to $se(3)$,\\ e.g. $\scriptsize
				\widehat{\boldsymbol{\xi}}=\left(  \begin{matrix}
					\widetilde{\boldsymbol{\kappa}}&\boldsymbol{\epsilon}\\
					\boldsymbol{0}&0
				\end{matrix}\right),\ \widehat{\boldsymbol{\eta}}=\left(  \begin{matrix}
					\widetilde{\boldsymbol{\omega}}&\boldsymbol{v}\\
					\boldsymbol{0}&0
				\end{matrix}\right)\in se(3) $ }}\\
		$(\cdot)^{\vee}$& --- & Mapping from $se(3)$ to $\mathbb{R}^6$.\\
		$\rho$ & kg/m$^3$ & Density of material. \\
		$R(s)$ & m & Cross-sectional radius. \\
		$E$ & Pa & Young's modulus. \\
		$\mu$ & --- & Coefficient of friction. \\
		$\mathbf{I}$& --- &identity matrix.\\
		$\boldsymbol{J}(s,t)$& --- & Jacobian matrix of Cosserat kinematics.\\
		\multirow{1}{*}{$\rm{ad}_{(\cdot)}$} & \multicolumn{1}{l}{\tabincell{l}{--- }}&\multicolumn{1}{l}{\tabincell{l}{The adjoint map of the Lie algebra, e.g.\\ $\scriptsize\rm{ad}_{\boldsymbol{\xi}}= \left(\begin{matrix}
					\widetilde{\boldsymbol{\kappa}}&\boldsymbol{0}_{3\times3}\\\widetilde{\boldsymbol{\epsilon}}&\widetilde{\boldsymbol{\kappa}}
				\end{matrix}\right)$, $\scriptsize\ \rm{ad}_{\boldsymbol{\eta}}= \left(\begin{matrix}
					\widetilde{\boldsymbol{\omega}}& \boldsymbol{0}_{3\times3}\\
					\widetilde{\boldsymbol{v}}&\widetilde{\boldsymbol{\omega}}
				\end{matrix}\right).$}}\\ 
		\multirow{1}{*}{$\boldsymbol{\mathcal{M}}$} & \multicolumn{1}{l}{\tabincell{l}{--- }}&\multicolumn{1}{l}{\tabincell{l}{Cross-sectional mass matrix.}}\\
		\multirow{1}{*}{${\rm{Ad}}_{\boldsymbol{g}(X)}$ } & \multicolumn{1}{l}{\tabincell{l}{--- }}&\multicolumn{1}{l}{\tabincell{l}{The matrix transforming the velocity or \\acceleration twist from body frame to \\inertial frame, \\i.e., $\scriptsize {\rm{Ad}}_{\boldsymbol{g}(X)}= \left(\begin{matrix}
					\boldsymbol{R}&\boldsymbol{0}_{3\times3}\\\widetilde{\boldsymbol{p}}\boldsymbol{R}&\boldsymbol{R}
				\end{matrix}\right)\in \mathbb{R}^{6\times6}.$}}\\	
	    \multirow{1}{*}{${\rm{Ad}}^*_{\boldsymbol{g}(X)}$ } & \multicolumn{1}{l}{\tabincell{l}{--- }}&\multicolumn{1}{l}{\tabincell{l}{The matrix transforming the wrench from\\contact frame to body frame, \\i.e., $\scriptsize {\rm{Ad}}^*_{\boldsymbol{g}(X)}= \left(\begin{matrix}
	    			\boldsymbol{R}&\widetilde{\boldsymbol{p}}\boldsymbol{R}\\\boldsymbol{0}_{3\times3}&\boldsymbol{R}
	    		\end{matrix}\right)\in \mathbb{R}^{6\times6}.$}}\\	
		$n$& --- & Number of sections divided for strain field.\\
		$m$& --- & Number of sections divided for contact field.\\	
		$\delta_n(s,t)$& m & Normal gap of contact in contact frame.\\
		$\boldsymbol{v}_t(s,t)$& m & \multicolumn{1}{l}{\tabincell{l}{Relative tangent velocity of contact in contact \\frame}}.\\
		$\Lambda_n(s,t)$& N/m & \multicolumn{1}{l}{\tabincell{l}{Normal contact load of collision in contact\\frame.}}\\
		$\boldsymbol{\Lambda}_t(s,t)$& N/m & \multicolumn{1}{l}{\tabincell{l}{Tangent contact load of collision in contact\\frame.}}\\
		$\boldsymbol{\Lambda}_c(s,t)$& N/m & Contact load of collision in contact frame.\\
		$\boldsymbol{\Lambda}_a(s,t)$& N/m & \multicolumn{1}{l}{\tabincell{l}{Contact load of articulated constraint in contact \\frame.}}\\
		${u}_n(t)$& --- & Slack variable of normal contact constraint.\\	
		$\boldsymbol{u}_t(t)$& --- & Slack variable of tangent contact constraint.\\
		$\boldsymbol{u}(t)$& --- & Slack variable of contact constraint.\\
		$\boldsymbol{\lambda}_c(t)$& --- & Assembly vector of slack variable $\boldsymbol{u}$.\\	
		$\boldsymbol{\lambda}_f(t)$& --- & \multicolumn{1}{l}{\tabincell{l}{Assembly vector of contact force of articulated \\constraints.}}\\	
		$\boldsymbol{\lambda}_a(t)$& --- & \multicolumn{1}{l}{\tabincell{l}{Assembly vector of contact force of fixed \\constraints.}}\\
		$\boldsymbol{G}_c(t)$& --- & \multicolumn{1}{l}{\tabincell{l}{Collision contact constraints.}}\\
		$\boldsymbol{G}_f(t)$& --- & \multicolumn{1}{l}{\tabincell{l}{Fixed contact constraints.}}\\
		$\boldsymbol{G}_a(t)$& --- & \multicolumn{1}{l}{\tabincell{l}{Articulated contact constraints.}}\\	
		\bottomrule
	\end{tabular}
	\label{notation_definitions}
\end{table}
\section{Problem statement}\label{sec::pro}
Due to the continuum nature of soft robot deriving a infinite DoF, it is difficult or impossible to get its exact dynamics handling contacts. Thus researchers usually use approximate technique to these models. Cosserat is a relative efficient way to approximate the dynamics of soft slender robot with a lower dimension than FEM on profiting the geometric nature of rod. Normally, if one can approximate the configuration of soft slender robot with a set of generalized coordinates $\boldsymbol{q}\in\mathbb{R}^n$, the dynamics without contacts could be described through the following second-order differential equations\cite{armanini2023soft}\cite{10027557}:
\begin{equation}\label{dyy}
	\boldsymbol{M}(\boldsymbol{q})\ddot{\boldsymbol{q}}+\boldsymbol{C}(\boldsymbol{q},\dot{\boldsymbol{q}})+\boldsymbol{K}\boldsymbol{q}=\boldsymbol{H}(\boldsymbol{q})\boldsymbol{\tau}+\boldsymbol{P}(\boldsymbol{q}),
\end{equation}
where $\boldsymbol{M}$ is the generalized mass matrix and $\boldsymbol{C}$ represents the contribution of viscosity and Coriolis acceleration. $\boldsymbol{K}$ is the stiffness matrix. $\boldsymbol{H}$ is the actuation matrix and $\boldsymbol{\tau}$ denotes the actuation force. $\boldsymbol{P}$ contains the contributions of the external force such as the gravity.
\begin{figure}[h]
	\centering
	\includegraphics[width=0.43\textwidth]{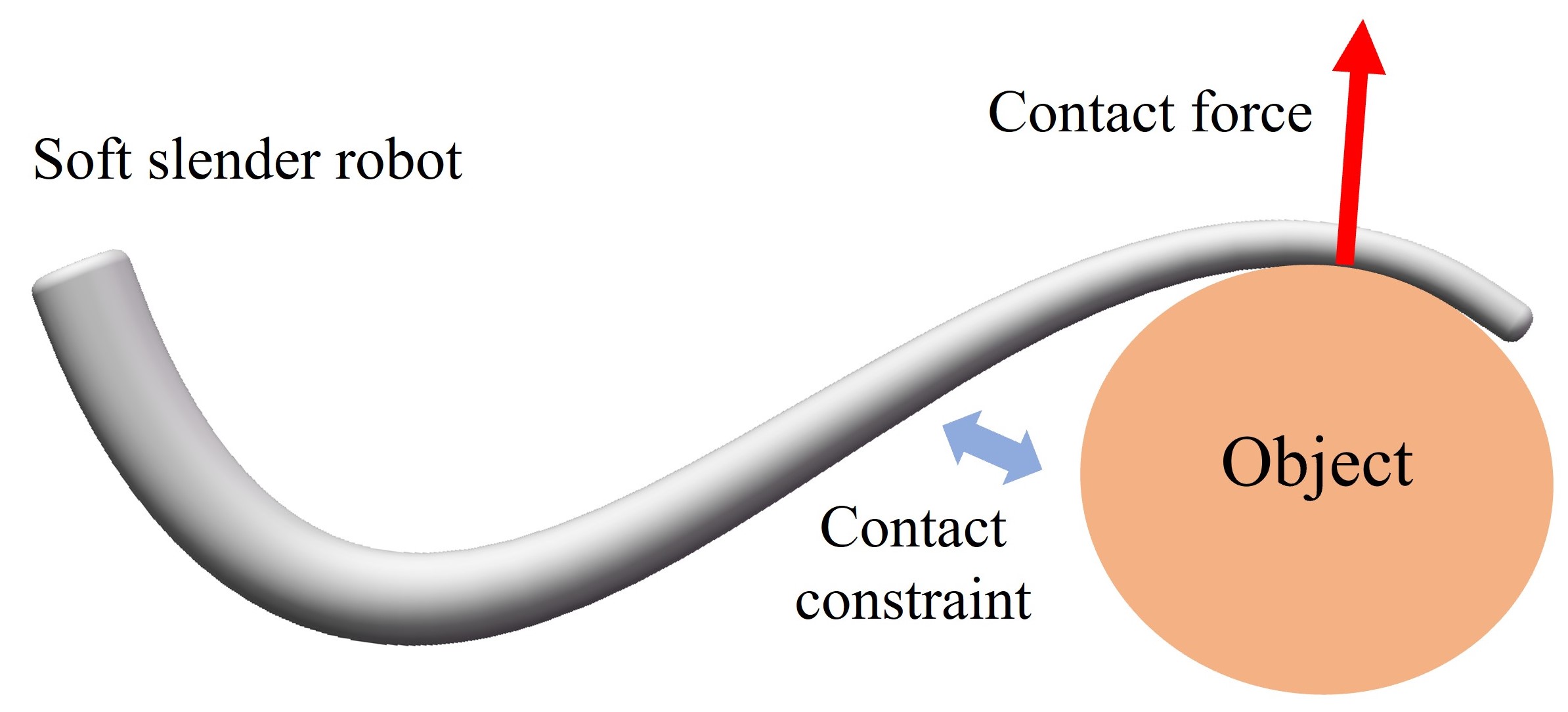}
	\caption{Soft slender robot contacting with object.}
	\label{fig:contact1}
\end{figure}
When contact exists (see Fig. \ref{fig:contact1}), the contribution of contact constraint to the dynamic system (\ref{dyy}) in generalized coordinates could be expressed by a set of Lagrange multipliers $\boldsymbol{\lambda}\in\mathbb{R}^m$ which physically represents the contact force. Note that the introduction of Lagrange multipliers introduces additional unknown variables to the system. Consequently, solving the system necessitates augmenting the equations with constraint equations that govern the contact behavior.  

On account of these backgrounds, we would like to highlight the following issues at the beginning of this paper:\\
(1). How to derive the dynamics of soft slender robot under contact with limited degree of freedoms without losing local accuracy? \\
(2). How to compute the contribution of the contact force?\\
(3). How to efficiently deal with the contact constraints of dynamics?

Those questions will be investaged in the following sections.
\section{Coordinates of soft slender robot based on Lie group}\label{sec::Kinematics}
Compared to traditional FEM, the Cosserat-based method can offer computational advantages for soft slender robot models. Due to its simplified geometric and kinematic description on Lie group, the Cosserat-based model can lead to reduced computational complexity and faster simulation, which bring advantage for robotic application.  In the following we will define the configuration as well as the surface of soft slender robot by different coordinates, which will be used hereafter to deduce the contact dynamics of soft slender robot.
\subsection{Backbone of soft slender robot}
\begin{figure}[h]
	\centering
	\includegraphics[width=0.48\textwidth]{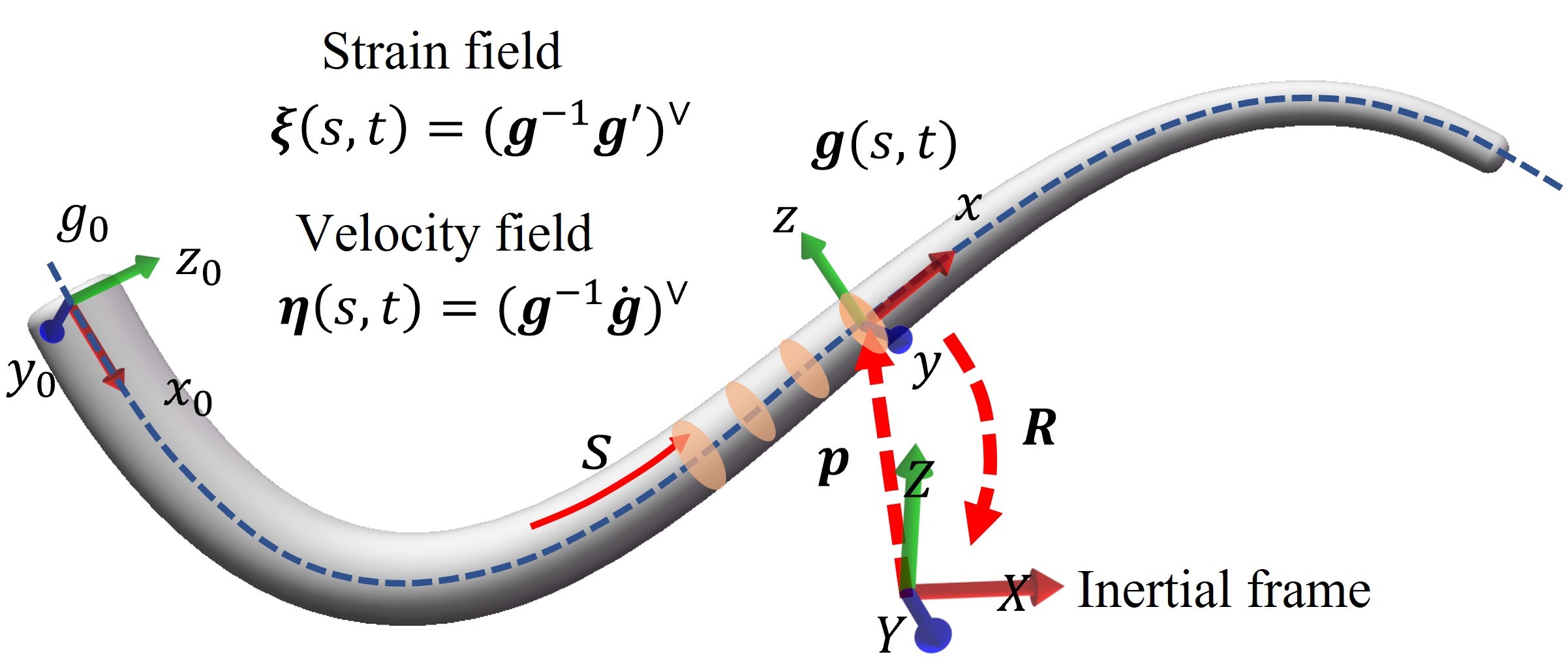}
	\caption{Backbone of the soft slender robot.}
	\label{fig:contact}
\end{figure}
In Cosserat rod theory, the soft slender robot is considered as a set of rigid cross-sections along its centerline, see Fig. \ref{fig:contact}.
The spatial configuration of each cross-sections is defined in $SE(3)$ and is represented by the continuous configuration tensor $\boldsymbol{g}(s,t)$, which describes the position and orientation of each cross-section with respect to the inertial (global) frame. $\boldsymbol{g}(s,t)$ is defined by rotation matrix $\boldsymbol{R}(s,t)\in \mathrm{SO}(3)$ and position vector $\mathbf{p}(s,t)\in \mathbb{R}^3$, which represent the orientation and position of each cross-section of the rod with respect to global frame, as depicted in Fig. \ref{fig:contact}. For a length normalized soft slender robot, its configuration space is defined as follows:
$$\mathcal{U}=\{\boldsymbol{g}: s \in[0,1] \mapsto \boldsymbol{g}(s,t) \in \mathrm{SE}(3)\},\ 	\boldsymbol{g}=\begin{bmatrix}
	\boldsymbol{R} & \boldsymbol{p} \\
	\boldsymbol{0}^{\top} & 1
\end{bmatrix}$$

The space variation of the configuration can be defined with the vector fields $\boldsymbol{\xi}(s,t)=(\boldsymbol{g}^{-1}\boldsymbol{g}^{\prime})^{\vee}$ which stand for the strain of the cross-section located at $s$ in body frame, with $\boldsymbol{\xi}(s,t)=[\boldsymbol{\kappa}^\top \ \boldsymbol{\epsilon}^\top]^\top \in \mathbb{R}^{6}$,
where $\boldsymbol{\kappa}(s,t)$ stands for the angular strain and $\boldsymbol{\epsilon}(s,t)$ represents the linear strain. By the above notations, the geometric model of soft slender robot is derived as below:
\begin{equation}\label{geom}
	\boldsymbol{g}^{\prime}=\boldsymbol{g} \widehat{\boldsymbol{\xi}}
\end{equation}
with initial condition $\boldsymbol{g}(0) = \boldsymbol{g}_0$.
As (\ref{geom}) indicates, if the initial configuration $\boldsymbol{g}_0$ and the overall strain space $\boldsymbol{\xi}(s)$ of the soft slender robot are known for an instant, the configuration space of the soft slender robot can be constructed by solving (\ref{geom}). Consequently, the configuration space $\mathcal{U}$ of the soft slender robot can be totally defined by a set of $\{\boldsymbol{g}_0, \boldsymbol{\xi}(s)\}$,
{defined as follows:}
\begin{equation}\label{key}
	\mathcal{U}=SE(3)\times\mathbb{S}
\end{equation}
where $SE(3)$ denotes the configuration space of $\boldsymbol{g}(0)$ and $\mathbb{S}$ denotes the strain field, with $\mathbb{S}=\{\boldsymbol{\xi}:s\in[0,1]\mapsto\boldsymbol{\xi}\in\mathbb{R}^6\}$.

As for velocity, the time variation of the configuration can be defined with the vector fields $\boldsymbol{\eta}=(\boldsymbol{g}^{-1}\dot{\boldsymbol{g}})^{\vee}$ which stand for the velocity twist of soft slender robot in local frame,
with
$
\boldsymbol{\eta}=[\boldsymbol{w}^\top \ \boldsymbol{v}^\top]^\top \in \mathbb{R}^{6}
$,
where $\boldsymbol{w}$ stands for the angular velocity and $\boldsymbol{v}$ represents the linear velocity with respect to local frame.
\subsection{Contact point on the surface of soft slender robot}
\begin{figure}[h]
	\centering
	\includegraphics[width=0.48 \textwidth]{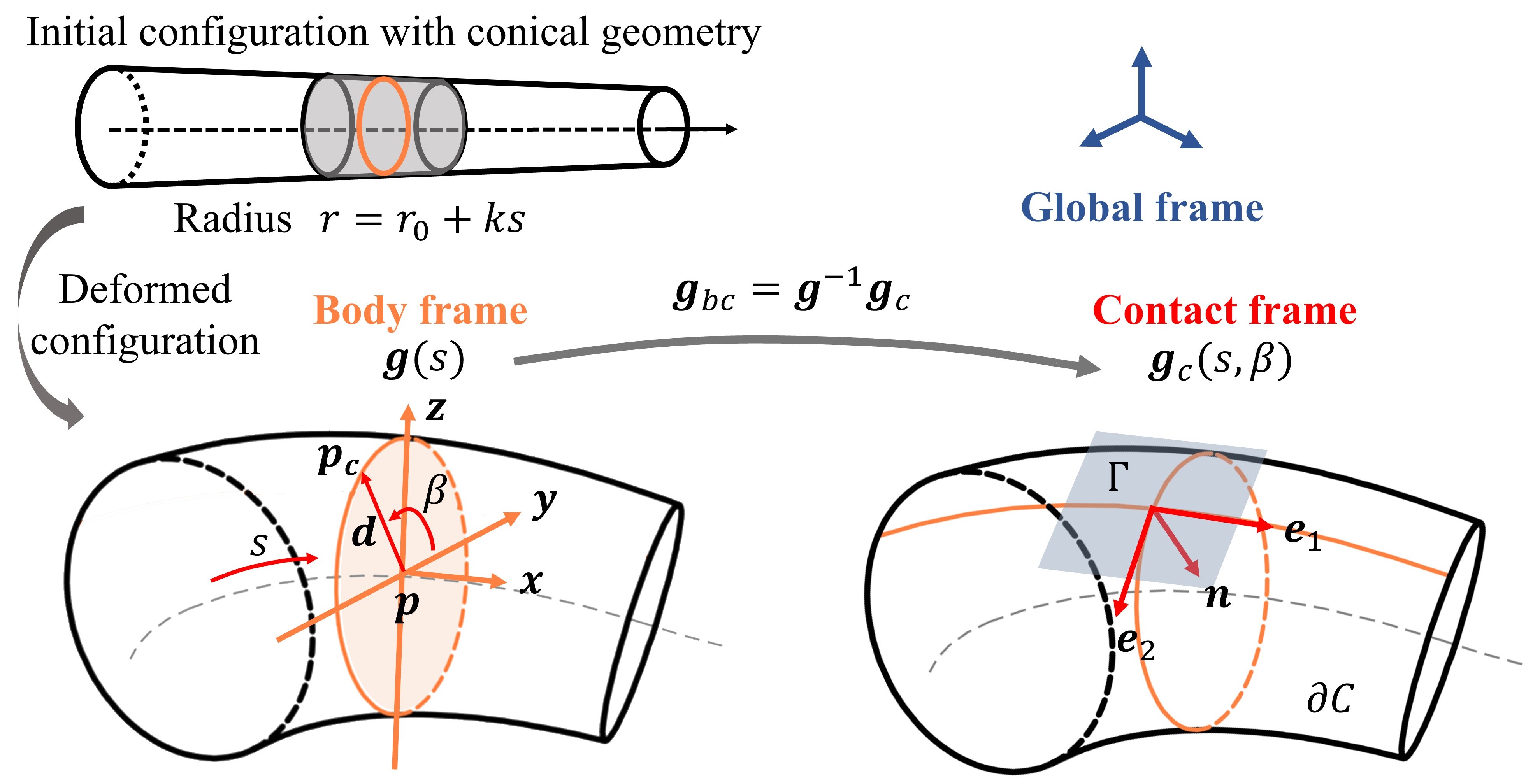}
	\caption{Coordinates of surface of soft slender robot.}
	\label{fig:surface}
\end{figure}

In the previous subsection, we focused on the geometric representation of the soft slender robot backbone by employing the coordinate $s$ to denote the position of each disc along its centerline. However, to describe the coordinates for every point on the surface $\partial\mathcal{C}$ of the soft slender robot, we introduce the second coordinate $\beta\in\mathbb{R}$. This new coordinate defines the position of a point on the disc relative to its center. As illustrated in Fig. \ref{fig:surface}, $\beta$ represents the rotation angle in radians of the contact point around the center point in the direction of $x$-axis. 
By using the coordinate $\boldsymbol{X} = [s,\beta]^\top$, we can precisely locate any point on the surface of the soft slender robot, which allows us to analyze the contact situation of the soft slender robot during various tasks and interactions. For any point on the surface, its position can be given by:
\begin{equation}\label{key}
	\boldsymbol{p}_c(s,\beta) = \boldsymbol{p}(s)+\boldsymbol{R}(s)\boldsymbol{d}(s,\beta)
\end{equation}
where $\boldsymbol{d}(s,\beta)$ is the local distance vector from the center-line of soft slender robot to the point on the surface, which is defined in the body frame. Since we assume that the form of cross-section never changes during the deformation of soft slender robot, this vector only depends on robot's original geometries of the soft slender robot. For example, for a conical soft slender robot with circular cross-section, as shown in Fig. \ref{fig:surface},
$\boldsymbol{d}(s,\beta)=\begin{bmatrix}
	0 & r\cos \beta & r\sin \beta
\end{bmatrix}^\top$, 
where $r=r_0+ks$ is the radius of the cross-section at length $s$. $r_0$ is the radius of the initial section and $k$ is the radial gradient with respect to $s$.
\subsection{Contact frame}
Once we define the contact point on the surface of slender robot, the contact frame at this point can be represented by a contact plane which is tangent to the surface of soft slender robot, as shown in Fig. \ref{fig:surface}. {This plane, noted as} $\Gamma$, can be defined by a pair of covariant tangent vectors $\{\boldsymbol{\tau}_1,\boldsymbol{\tau}_2\}$, with:
$$\begin{aligned}
	\boldsymbol{\tau}_1 =\frac{\partial \boldsymbol{p}_c}{\partial s}=\frac{\partial\boldsymbol{p}}{\partial s}+\frac{\partial\boldsymbol{R}}{\partial s}\boldsymbol{d}+\boldsymbol{R}\frac{\partial\boldsymbol{d}}{\partial s}
	=\boldsymbol{R}(\boldsymbol{\epsilon}+\tilde{\boldsymbol{\kappa}}\boldsymbol{d}+\frac{\partial\boldsymbol{d}}{\partial s})
\end{aligned}$$ 
$$\begin{aligned}
	\boldsymbol{\tau}_2 = &\frac{\partial \boldsymbol{p}_s}{\partial \beta}=\frac{\partial\boldsymbol{d}}{\partial \beta}
\end{aligned}$$ 
The unit normal vector of this tangent surface is then given by:
\begin{equation}\label{cor}
	\boldsymbol{n}=\frac{\boldsymbol{\tau}_1\times\boldsymbol{\tau}_2}{\Vert\boldsymbol{\tau}_1\times\boldsymbol{\tau}_2\Vert}
\end{equation}
Note that the basis $\{\boldsymbol{\tau}_1,\boldsymbol{\tau}_2,\boldsymbol{n}\}$ is not necessarily orthogonal in the deformed configuration, i.e., $\boldsymbol{\tau}_1^\top\boldsymbol{\tau}_2\neq0$, thus a new orthogonal basis can be defined from $\{\boldsymbol{\tau}_1,\boldsymbol{\tau}_2,\boldsymbol{n}\}$ by replacing $\boldsymbol{\tau}_1$ and $\boldsymbol{\tau}_2$ with:
$$\boldsymbol{e}_1=\frac{\boldsymbol{\tau}_1}{\Vert\boldsymbol{\tau}_1\Vert}\ , \ \boldsymbol{e}_2=	\frac{\boldsymbol{n}\times\boldsymbol{\tau}_1}{\boldsymbol{n}\times\Vert\boldsymbol{\tau}_1\Vert}$$
We can finally use the orthogonal basis $\{\boldsymbol{e}_1,\boldsymbol{e}_2,\boldsymbol{n}\}$ to define the contact frame.
{Note that these three unit vectors are all defined in the global frame, thus can be regarded as a rotation matrix} noted as $\boldsymbol{R}_c$ which transfers the contact frame to global frame:
$$\boldsymbol{R}_c=\begin{bmatrix}
	\boldsymbol{n}&\boldsymbol{e}_1&\boldsymbol{e}_2
\end{bmatrix}$$
Based on this matrix, we can then define a configuration tensor $\boldsymbol{g}_c\in SE(3)$ which represents the position and orientation of the contact frame with respect to the global frame.
$$\boldsymbol{g}_c=\begin{bmatrix}
	\boldsymbol{R}_c& \boldsymbol{p}_c\\
	\boldsymbol{0}^\top&1
\end{bmatrix}$$
\section{Contact force and contact constraint}\label{sec::cont_constrain}
\begin{figure}[h]
	\centering
	\includegraphics[width=0.4\textwidth]{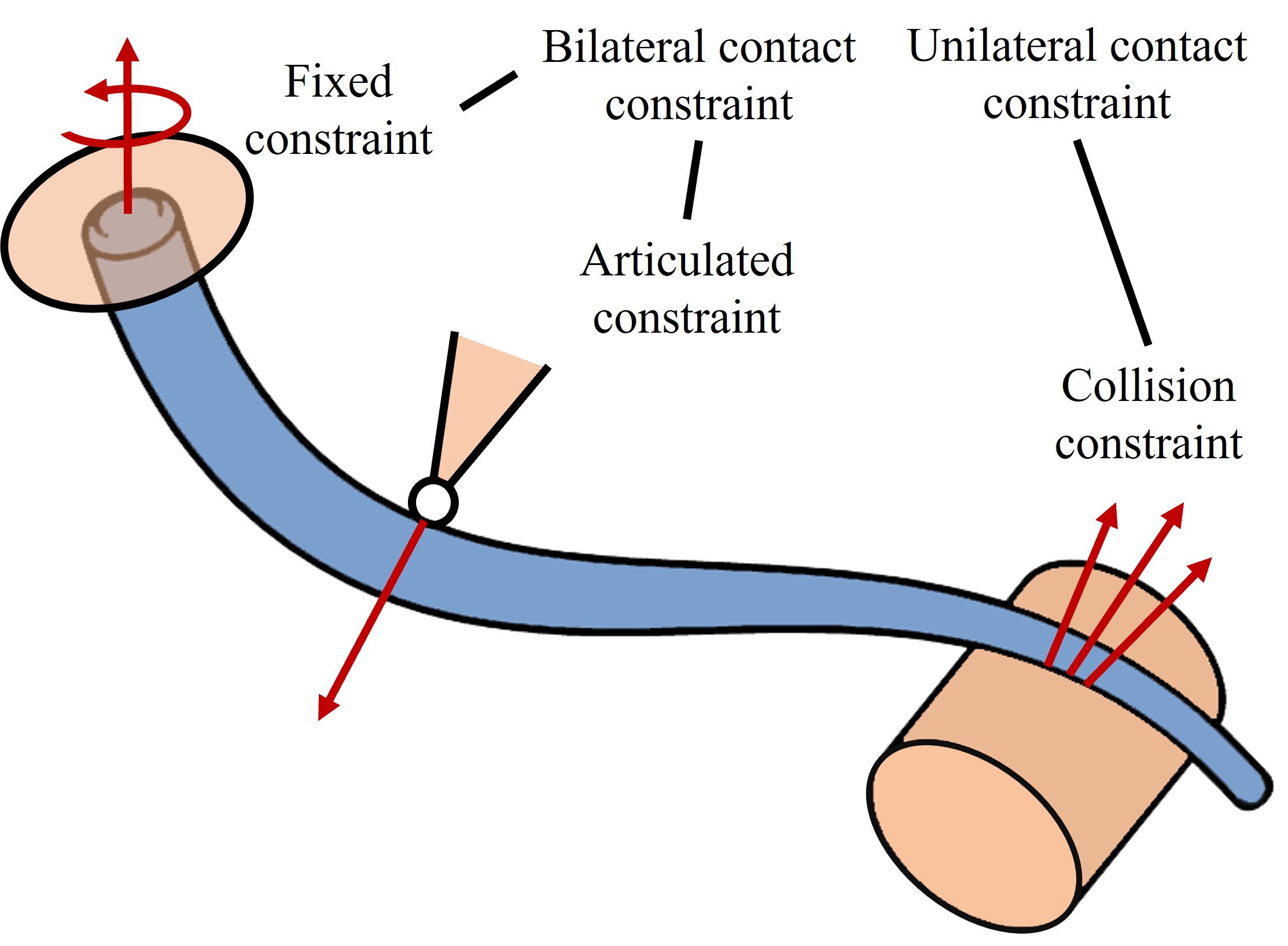}
	\caption{Different contact constraints of soft slender robot.}
	\label{fig:type}
\end{figure}

As depicted in the Fig. \ref{fig:type}, contact constraints can arise in soft robotics, including fixed constraints, articulated constraints, and collision constraints. Among these, fixed constraints and articulated constraints can be represented using {equalities} and fall into the category of bilateral constraints. On the other hand, collision constraints are typically expressed using inequalities and fall into the category of unilateral constraints{\cite{johnson1987contact}}.
\subsection{Contact force}
All contact constraints give rise to contact forces at the contact points. For any contact frame, we denote the contact force expressed in this {contact} frame as ${\boldsymbol{\Lambda}}_c\in \mathbb{R}^6$. Then, we denote ${\boldsymbol{\Lambda}}_c^*$ as the equivalent wrench with respect of body frame, which is generated by translating the distributed contact force onto the center of cross-section. There exists the following relationship between them: 
\begin{equation}\label{force}
	{\boldsymbol{\Lambda}}_c^*=\mathrm{Ad}^*_{\boldsymbol{g}_{bc}}{\boldsymbol{\Lambda}}_c
\end{equation}
where the coadjoint representation of the Lie group $\mathrm{Ad}^*$ allows changing the wrench from contact frame to body frame, with $\boldsymbol{g}_{bc}=\boldsymbol{g}^{-1}\boldsymbol{g}_c$, the configuration tensor of contact frame with respect to body frame. 
\begin{figure}[h]
	\centering
	\includegraphics[width=0.48\textwidth]{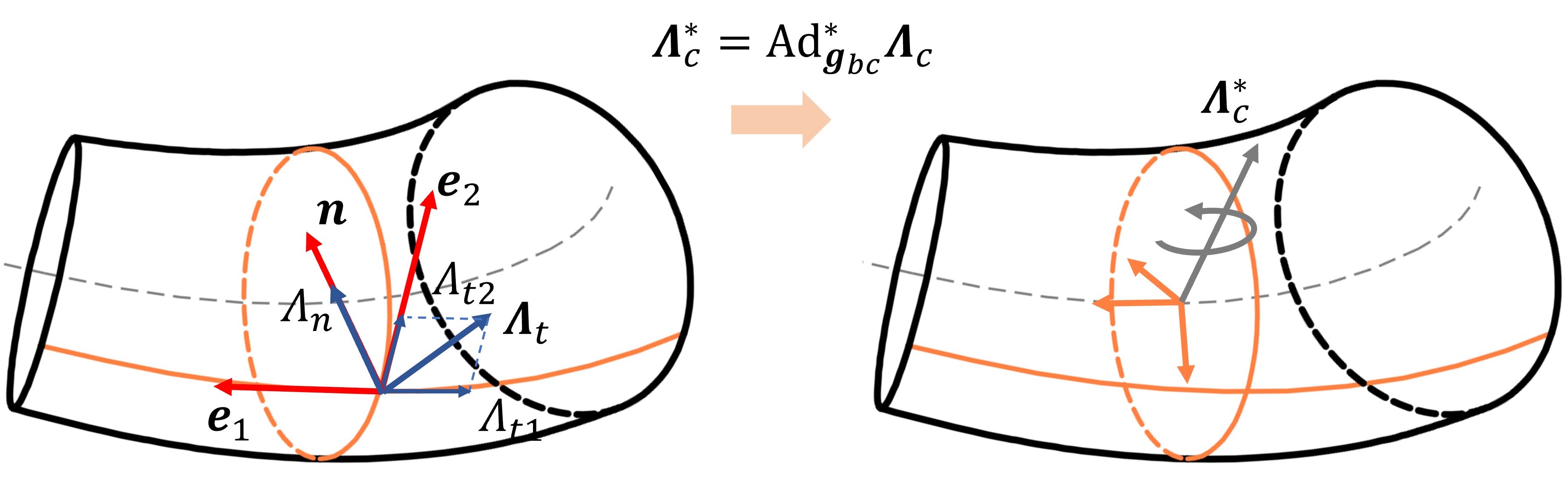}
	\caption{Transformation of contact load from contact point to center line.}
	\label{fig:contacttrans}
\end{figure}
\subsection{Bilateral contact constrain}\label{sec::Bilateral_control_constrain}
Bilateral constraints are mathematical representations of kinematic pairs, such as spherical, or revolving joints. These constraints can be expressed as algebraic equations that impose restrictions on the relative positions of two bodies.
\subsubsection{Fixed constraint}
In engineering applications or some simulation scenarios, some parts of the soft slender robot will be totally fixed by the external environment. For example, the head section of the soft slender robot is fixed on a mobile base or {some of its} parts are hinged to external objects like truss system\cite{9442856}. These persistent constraints can be regarded as bilateral constraints. Specifically, the satisfaction of bilateral constraints can be realized by adding Lagrangian multipliers in the system. For any points $s\in[0,1]$ along the centerline of the soft slender robot, if we want to fix some sections of soft slender robot to the desired configuration, that is, given a constrained configuration matrix {$\boldsymbol{g}_{f}(s)$}, then the bilateral constraint can be expressed as:
$${\boldsymbol{G}_{f}(s)}\in \mathbb{R}^6=
\left(\log(\boldsymbol{g}_{f}^{-1}\boldsymbol{g})\right)^\vee=\mathbf{0}$$
which is equivalent to the following bilateral constriant $\boldsymbol{g}_{f}(s) - \boldsymbol{g}(s) = \boldsymbol{0}$, stated in the matrix form.
\subsubsection{Articulated constraint}
Articulatd constraints in soft slender robots impose limitations on the translation of contact points while permitting rotational freedom at the cross-sections of these points. Consequently, only the three translational degrees of freedom are constrained by articulated constraints. For $s\in[0,1]$, denoting the desired position of the joint as {$\boldsymbol{p}_{a}(s)$ with respect to the global frame}, the articulated constraint is given by the following equation:
$$\mathbb{R}^3\ni\boldsymbol{G}_{a}(s)
=\boldsymbol{p}_{a}-\mathbf{U}\boldsymbol{g}_c(s,\beta)\mathbf{A}=\mathbf{0}$$
where $\mathbf{U}=[1\ 1\ 1\ 0]$ and {$\mathbf{A}=[0\ 0\ 0\ 1]^\top$}.
Note that for the articulated constraint, the joint can only generate the force but not the moment, which means that $\boldsymbol{\Lambda}_c$ is not full filled. Denoting the generated force as {$\boldsymbol{\Lambda}_a\in \mathbb{R}^3$}, (\ref{force}) should be rewritten as:
$${\boldsymbol{\Lambda}}_c^*=\mathrm{Ad}^*_{\boldsymbol{g}_{bc}}\mathbf{B}^\top{\boldsymbol{\Lambda}}_{a}$$ 
where $\mathbf{B}=[\mathbf{0}_{3\times3} \ , \mathbf{I}_{3\times3}]$.
\subsection{Unilateral contact constrain}\label{sec::conact_constrain}
In the field of contact mechanics, the unilateral constraint refers to a mechanical constraint that prohibits any form of penetration or interpenetration between two bodies, whether they are rigid or flexible. This constraint ensures that the tangential contact forces between the bodies adhere to the contact law, which governs the interaction between contacting surfaces.

For collision constraint, the generated force is composed by the normal contact force and tangent friction force, i.e.,
${\boldsymbol{\Lambda}}_c=[\boldsymbol{0}_{1\times3} , {\Lambda}_n  , {\Lambda}_{t1}, {\Lambda}_{t2}]^\top$, 
{where $\Lambda_n$, $\Lambda_{t1}$ and $\Lambda_{t2}$} denote respectively the normal contact force in the direction of $\boldsymbol{n}$, the friction force in the direction of $\boldsymbol{e}_1$ and $\boldsymbol{e}_2$. Thus (\ref{force}) should be rewritten as:
\begin{equation}\label{125}
	{{\boldsymbol{\Lambda}}_c^*}=\mathrm{Ad}^*_{\boldsymbol{g}_{bc}}(\mathbf{B}_n^\top \Lambda_n+\mathbf{B}_t^\top\boldsymbol{\Lambda}_t)
\end{equation}
where $\mathbf{B}_n=[\mathbf{0}_{1\times3} , 1 , \mathbf{0}_{1\times2}]$, $\mathbf{B}_t=[\mathbf{0}_{2\times4} , \mathbf{I}_{2\times2}]$ and $\boldsymbol{\Lambda}_t=[{\Lambda}_{t1}, {\Lambda}_{t2}]^\top$.

In accordance with the problem statement, the collision constraints encompass a set of inequalities that involve contact gaps and contact sliding velocities. Hence, our initial objective is to elucidate how to define contact gaps and contact sliding velocities within the framework of Cosserat rod theory.

\begin{figure}[h]
	\centering
	\includegraphics[width=0.35\textwidth]{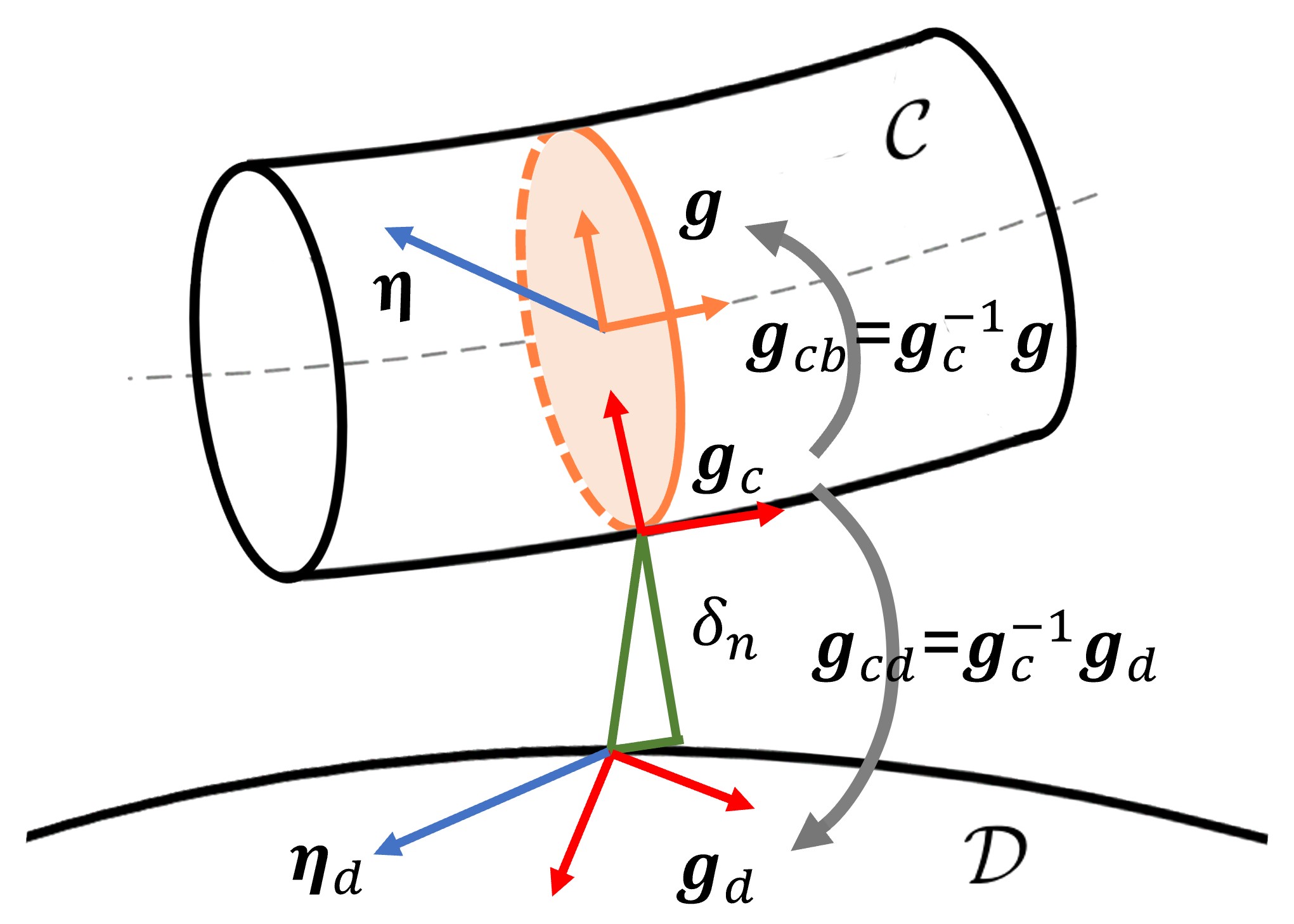}
	\caption{Geometries of the contact pair.}
	\label{fig:2contact}
\end{figure}

As shown in Fig. \ref{fig:2contact}, we designate the contact frame of the soft slender robot as $\boldsymbol{g}_c$, serving as the master contact frame. We also denote a known slave contact frame $\boldsymbol{g}_d$, located on the surface $\partial\mathcal{D}$ of another body $\mathcal{D}$. These two frames constitute a contact pair. In order to represent the relative position and orientation of slave contact frame on $\partial\mathcal{D}$ with respect to contact frame $\boldsymbol{g}_c$ on $\partial\mathcal{C}$, we employ a tensor denoted as $\boldsymbol{g}_{cd}=\boldsymbol{g}_c^{-1}\boldsymbol{g}_d$.

The normal gap is the relative position vector projected on the opposite direction of normal vector $\boldsymbol{n}$ and can be obtained from the relative configuration tensor of {contact via the following relation:}.
\begin{equation}\label{key}
	\delta_n=-\mathbf{P}\boldsymbol{g}_{cd}\mathbf{A}
\end{equation}
with $\mathbf{P}=[1\ 0\ 0\ 0]$ and $\mathbf{A}=[0\ 0\ 0\ 1]^\top$.

The normal force and normal gap should always be non-negative, i.e.,
$\delta_n\geq0$ , $ \Lambda_n \geq0$.
This principle imposes a constraint on the occurrence of mutual penetration among objects in contact, thereby prohibiting their ability to traverse another one. The magnitude of the normal contact force is invariably non-negative, and contact forces are solely generated upon contact establishment.

The friction force is constrained within a convex set $\mathcal{C}_f$ corresponding to the friction cone, such that $\boldsymbol{\Lambda}_{t}\in \mathcal{C}_f$. In our work we use the Coulomb's law to model the friction. The section of the Coulomb's friction cone, i.e., the disk $\mathcal{D}(\mu \Lambda_{n})$ is defined by
\begin{equation}\label{friction cone}
	\mathcal{D}(\mu \Lambda_{n})=\{\Lambda_{t}|\mu \Lambda_{n}-\Lambda_{t}\geq0\}
\end{equation}
with $\mu$ being the coefficient of friction. 

Denoting the velocity of backbone of soft slender robot with respect to body frame as $\boldsymbol{\eta}$ and supposing the velocity of the slave contact point on object $\mathcal{D}$ with respect to the slave contact frame $\boldsymbol{g}_d$ is $\boldsymbol{\eta}_d$. The slip velocity is defined as the projection of the relative contact velocity onto the tangential plane at the contact point.
Thus, the relative velocity twist of contact point with respect to contact frame $\boldsymbol{g}_c$ of $\partial\mathcal{C}$ can be then given by:
\begin{equation}\label{vv}
	\boldsymbol{\eta}_c=\mathrm{Ad}_{\boldsymbol{g}_{cb}}\boldsymbol{\eta}-\mathrm{Ad}_{\boldsymbol{g}_{cd}}\boldsymbol{\eta}_d
\end{equation}  
where $\boldsymbol{g}_{cb}=\boldsymbol{g}_c^{-1}\boldsymbol{g}$ and the adjoint representation of the Lie group $\mathrm{Ad}_{\boldsymbol{g}_i^{-1}\boldsymbol{g}_j}$ allows changing the velocity twist from frame j to frame i.
{Consequently} the slip tangent velocity $\boldsymbol{v}_{t}\in \mathbb{R}^2$ on $\partial\mathcal{C}$ with respect to contact frame equals to:
\begin{equation}\label{slip}
	\boldsymbol{v}_{t}=\mathbf{B}_t\boldsymbol{\eta}_c
\end{equation} 
{where $\mathbf{B}_t$ was defined in (\ref{125}).}

The Coulomb's friction cone contains two different states (stick $v_t=0$ or slide $v_t\neq0$) of a contact point which is given by:
\begin{subnumcases}{\label{stick or slide}}
	\mathrm{Stick:}\   \mu\Lambda_{n}-\Vert\boldsymbol{\Lambda}_{t}\Vert>0\ ,\ \boldsymbol{v}_t=\boldsymbol{0} \label{stick}\\		
	\mathrm{Silde:} \  \boldsymbol{\Lambda}_{t}=-\mu \Lambda_{n}\mathbf{T} \ ,\ \boldsymbol{v}_t\neq\boldsymbol{0} \label{slide}
\end{subnumcases}
Here $\mathbf{T}$ is an identity vector which indicates the tangential sliding direction with $\mathbf{T}=\frac{\boldsymbol{v}_t}{\Vert \boldsymbol{v}_t\Vert}$.

There are two main approaches to model the above unilateral constraints. The first is based on the regularization of contact constraint such as penalty methods, while the second is based on non-smooth contact dynamics, modeling the contact constraint as the exact inequalities.

Within penalty methods, the constraint is characterized as soft, allowing for the possibility  of penetration between two contacting bodies. In the event of geometric penetration between these bodies, a penalty potential energy term is incorporated into the studied system, wherein its magnitude is directly proportional to the degree of penetration. The penalty potential will generate the resisted forces in normal and tangential directions.
Penalty formulations in constraint-based simulations offer the advantage of being unconstrained and relatively straightforward to simulate and differentiate. They are particularly useful when it comes to stability, as implicit integration methods allow for stable simulations across a wide range of stiffness values. However, one limitation of penalty methods is that they allow for small constraint violations. While minor violations of normal constraints may not be visually apparent, softening the static friction constraint can lead to undesirable artifacts in certain scenarios. For example, when simulating a heavy object resting on an inclined surface under static friction, if the simulation runs for a long duration, the introduction of tangential slipping due to softened stick constraints can eventually become visually noticeable. Therefore, in our work, we did not utilize this method, but instead, we consistently treated contact as a hard constraint throughout, which will be introduced in the next subsection.
\subsection{NCP formulation} 
As discussed in the previous subsection, the precise approach involves formulating the contact as a hard constraint. In the case of unilateral constraints, this formulation is commonly expressed as nonlinear complementary conditions.  
\subsubsection{Normal contact constraint} 
the NCP of normal contact is given by the Signorini's condition\cite{wriggers2006computational}:
\begin{equation}\label{nncp}
	{0}\leq \delta_n\ \bot\ {\Lambda}_{n}\geq{0}
\end{equation}
where the symbol "$\bot$" denotes that the product of $\delta_n$ and ${\Lambda}_{n}$ equals to zero.
\subsubsection{Tangential contact constraint}
The tangential contact constraint is about friction. We first denote $\boldsymbol{\Lambda}_{t}\in\mathbb{R}^2$ as the friction force of any contact point in contact frame on the {surface} $\partial \mathcal{C}$. $\boldsymbol{v}_{t}\in\mathbb{R}^2$ stands for the slip velocity.
{Then according to the maximal dissipation principle\cite{drumwright2011modeling}}, the instantaneous power of friction force at this contact point given by $\boldsymbol{\Lambda}_{t}^\top\boldsymbol{v}_{t}$ should always takes the minimum within the limit of friction cone, i.e., $\boldsymbol{\Lambda}_{t}$ should satisfy the following minimization:
$$\begin{aligned}
	&\ \ \boldsymbol{\Lambda}_{t}=\mathrm{argmin}\ \ \ \boldsymbol{\Lambda}_{t}^\top\boldsymbol{v}_{t}\\
	&\mathrm{subject\ to} \ \ \  \mu\Lambda_{n}-\Vert\boldsymbol{\Lambda_{t}}\Vert\geq0
\end{aligned}$$ 
where $\mu$ is the coefficient of friction. The first order Karush-Kuhn-Tucker conditions of above minimization is then given by:  
\begin{equation}\label{c22}
	\boldsymbol{v}_{t}+\lambda_{{sl}}\nabla{\Vert\boldsymbol{\Lambda}_{t}\Vert}=\mathbf{0}
\end{equation}
\begin{equation}\label{c33}
	0\leq\lambda_{{sl}}\ \bot\ \mu\Lambda_{n}-\Vert\boldsymbol{\Lambda}_{t}\Vert\geq0
\end{equation}
with $\lambda_{{sl}}$ the Lagrange multiplier. In fact, the above two equations can be further simplified. First, one can derive $\nabla{\Vert\boldsymbol{\Lambda}_{t}\Vert}=\boldsymbol{\Lambda}_{t}/\Vert\boldsymbol{\Lambda}_{t}\Vert$. Subsequently, taking this equation into (\ref{c22}) we can get $\boldsymbol{v}_{t}=-\lambda_{{sl}}\boldsymbol{\Lambda}_{t}/\Vert\boldsymbol{\Lambda}_{t}\Vert$. Then taking the norm of both two sides derives $\lambda_{{sl}}={\Vert \boldsymbol{v}_{t}\Vert}$. 
Finally, (\ref{c22}) and (\ref{c33}) are transferred to the following complementary conditions where the Lagrange multiplier is eliminated:
\begin{equation}\label{c222}
	\Vert\boldsymbol{\Lambda}_{t}\Vert\boldsymbol{v}_{t}+\Vert \boldsymbol{v}_{t}\Vert\boldsymbol{\Lambda}_{t}=\mathbf{0}
\end{equation}
\begin{equation}\label{c333}
	0\leq\Vert \boldsymbol{v}_{t}\Vert \ \bot\ \mu\Lambda_{n}-\Vert\boldsymbol{\Lambda}_{t}\Vert\geq0
\end{equation}
However, {it is clear that} the coupling between normal and frictional complementarity is non-convex and non-smooth. In the next section, we propose a smoothing method to solve this issue.
\section{Reformulation of the contact constraint}\label{Smoothing}
Mathematically, the nonlinear complementary constraint (\ref{nncp}) can be linearized to form a linear complementary problem (LCP).
However, the tangential constraints (\ref{c222}) and (\ref{c333}) cannot be directly formulated as an LCP due to the nonlinearity of the friction cone. A common approach is to approximate the Coulomb's friction cone with a pyramid. This allows for the utilization of LCP solvers such as {Lemke's algorithm\cite{stewart1996implicit} and Gauss-Seidel method\cite{duriez2005realistic}}. Despite some advancements made in these methods specifically for simulating contact dynamics, the computational efficiency remains a challenge. 
In order to ensure both accuracy and efficiency, we propose a novel approach to reformulate the contact constraint as a smooth equality constraint. This new formulation aims to strike a balance between computational accuracy and efficiency.
\subsection{NCP-function and slack variable}
The idea is to convert the NCP into a nonlinear function (NCP-function) in order to formulate the NCP as the equality constraint. Since the work by Mangasarian \cite{mangasarian1976equivalence} it has been well known that by means of a suitable function $\phi:\mathbb{R}^2\rightarrow\mathbb{R}$, the complementary condition $a\geq0,b\geq0,ab=0$ can be transferred to an equivalent nonlinear equation: $\phi(a,b)=0$. This technique has drawn interest for solving complementary programming in the field of mathematics in these years. Our work is mainly based on this idea, using a slack variable to define both $a$ and $b$.

Denoting $u\in\mathbb{R}$ as the slack variable and $D(u)$ as the Heaviside function with $D(u)=(\mathrm{sgn}(u)+1)/2$. The complementary condition of $a,b$ is equivalent to the following definition:
\begin{equation}\label{aa}
	a=D(u)u \ , \ b=D^*(u)u
\end{equation}
with $D^*(u)=-D(-u)$. 
%
It is evident that with the aforementioned definitions, the complementary condition holds for any $u\in\mathbb{R}$. Consequently, the nonlinear complementary constraint is now transformed into a manifold defined by (\ref{aa}) with respect to the slack variable $u$.
\subsection{Reformulation of collision constraint}
\subsubsection{Normal contact}
Returning to the topic of normal contact, we can utilize a variable $u_n\in \mathbb{R}$ to represent both the normal contact force $\Lambda_n$ and the normal gap $\delta_n$ at each contact point as the following definitions:
\begin{equation}\label{nf}
	\Lambda_n=D(u_n)u_n \ , \ \delta_n=D^*(u_n)u_n
\end{equation}
\subsubsection{Friction contact}
The equality constraint (\ref{c222}) means that $\boldsymbol{v}_{t}$ and $\boldsymbol{\Lambda}_{t}$ are always parallel and opposite, it is possible to define them by the parametric equation. At this aim, we first represent them in the polar coordinates, i.e., (\ref{c222}) equals to the following representation: 
\begin{equation}\label{polar}
	\boldsymbol{v}_{t}=
	\rho_1
	\begin{bmatrix}
		\sin\theta\\ \cos\theta
	\end{bmatrix}\ , \ 
	\boldsymbol{\Lambda}_{t}=-\rho_2
	\begin{bmatrix}
		\sin\theta\\ \cos\theta
	\end{bmatrix}
\end{equation}
where $\rho_1\geq 0$ and $\rho_2\geq 0$. Therefore, we can rewrite the complemantarity (\ref{c333}) as:
$$0\leq \rho_1 \ \bot\ \mu\Lambda_{n}-\rho_2\geq0$$
which is equivalent to: $\forall u_t>0$,
\begin{equation}\label{plus}
	\rho_1=f(u_t-\mu\Lambda_{n}) \ ,\ \rho_2=\mu\Lambda_{n}-f(\mu\Lambda_{n}-u_t)
\end{equation}
where function $f(x)$ denotes $D(x)x$.
Then by taking the above representation into (\ref{polar}) one can get: $\forall u_t>0$,
\begin{equation}\label{polar1}
	\boldsymbol{v}_{t}=
	f(1-\frac{\mu\Lambda_{n}}{u_t})
	\begin{bmatrix}
		u_t\sin\theta\\ u_t\cos\theta
	\end{bmatrix}
\end{equation}
\begin{equation}\label{polar2}
	\boldsymbol{\Lambda}_{t}=\left(-\frac{\mu\Lambda_{n}}{u_t}+
	f(\frac{\mu\Lambda_{n}}{u_t}-1)\right)
	\begin{bmatrix}
		u_t\sin\theta\\ u_t\cos\theta
	\end{bmatrix}
\end{equation}
Note that the vector part 
$\begin{bmatrix}
	u_t\sin\theta & u_t\cos\theta
\end{bmatrix}^\top$ represents an arbitrary vector defined in the polar coordinates, thus it can be rewritten in the Cartesian coordinates as $\boldsymbol{u}_t\in \mathbb{R}^2$ with $u_t=\Vert\boldsymbol{u}_t\Vert$. By doing so, we can describe (\ref{polar1}) and (\ref{polar2}) again in the Cartesian coordinates:
\begin{equation}\label{polar11}
	\boldsymbol{\Lambda}_{t}=W(\boldsymbol{u}_t){\boldsymbol{u}_t}\ , \ 
	\boldsymbol{v}_{t}=W^*(\boldsymbol{u}_t){\boldsymbol{u}_t} 
\end{equation}
where $$W(\boldsymbol{u}_t)=-\frac{\mu\Lambda_{n}}{u_t}+f(\frac{\mu\Lambda_{n}}{u_t}-1)\ , \
W^*(\boldsymbol{u}_t)=f(1-\frac{\mu\Lambda_{n}}{u_t}) 
$$
\begin{remark}
	Note that (\ref{polar11}) has no definition when $\boldsymbol{u}_t=\mathbf{0}_{2\times1}$, thus we define $W(0,0)=W^*(0,0)=\mathbf{0}_{2\times1}$ and $\nabla W(0,0)=\frac{\mu\Lambda_{n}}{l}\mathbf{I}_{2\times2}$, $\nabla W^*(0,0)=\mathbf{0}_{2\times2}$. 
\end{remark}

{For convenience purposes, we define the slack vector $\boldsymbol{u}=[{u}_n \ \boldsymbol{u}_t^\top]^\top$ to cover both normal constraint and friction constraint. Based on the previous expression of contact forces, we can establish the relationship between the relaxation vector and contact forces using the following equation.
\begin{equation}\label{sss}
	\boldsymbol{\Lambda}_c=(\mathbf{B}_n^\top D\mathbf{C}_n+\mathbf{B}_t^\top W\mathbf{C}_t)\boldsymbol{u}
\end{equation}
where $\mathbf{C}_n=[1 \ \mathbf{0}_{1\times2}]$ and $\mathbf{C}_t=[\mathbf{0}_{2\times1} \ \mathbf{I}_{2\times2}]$. $\mathbf{B}_n$ and $\mathbf{B}_t$ was defined in (\ref{125}).
Similarly, the constraints of normal gap and the slip tangent velocity are written as:
\begin{equation}\label{lll}
	\delta_n=D^*\mathbf{C}_n\boldsymbol{u}\ , \ \boldsymbol{v}_{t}=W^*\mathbf{C}_t\boldsymbol{u}
\end{equation}}
\subsection{Smooth reformulation}\label{smoothfunc}
Nevertheless, the aforementioned formulation includes the Heaviside function, which is known to be discontinuous. Additionally, the function $f(x)$ in (\ref{plus}) possesses $C^0$ continuity, meaning that it is only continuous but not necessarily to be smooth. As a consequence, during the numerical solution process of dynamics, the presence of these functions leads to sudden changes, requiring additional iterations to achieve a solution. To address this concern, we propose a smooth approximation in our work, whereby the Heaviside function in (\ref{aa}) is replaced {by some well-known smooth functions\cite{mangasarian1994class}, which will be discussed hereafter.}
\subsubsection{Sigmoid function}
\begin{figure}[h]
	\centering
	\includegraphics[width=0.52\textwidth]{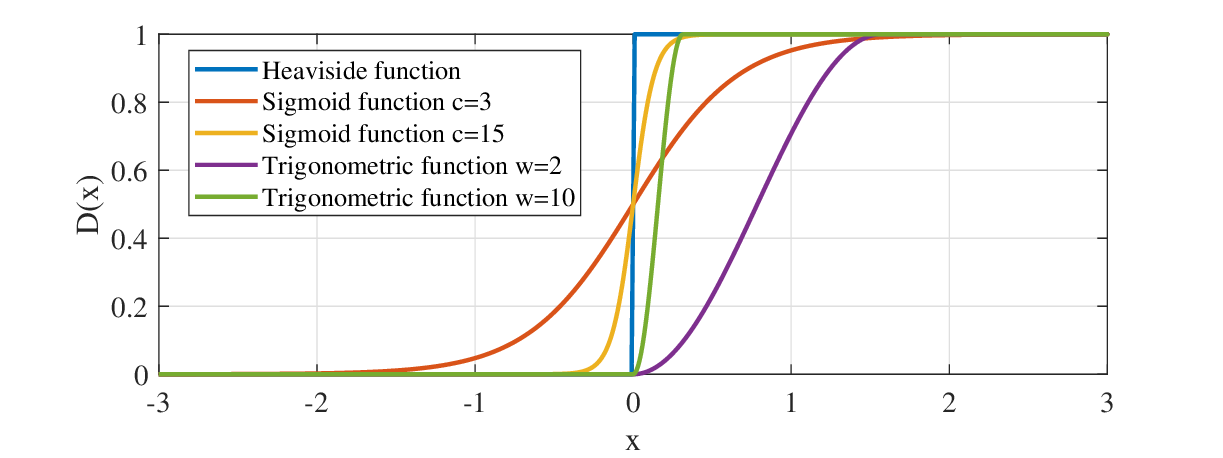}
	\caption{Proposed function for the reformulation of complementary constraint.}
	\label{fig:normal_contact1}
\end{figure}

The first smooth function we proposed is based on the sigmoid function defined as follows:
\begin{equation}\label{zzzz}
	D(x)=\frac{1}{1+e^{-cx}} \ , \ c>0
\end{equation}
The parameter $c$ plays a crucial role in determining the level of smoothing in (\ref{zzzz}). As $c$ approaches infinity, (\ref{zzzz}) progressively converges to the Heaviside function, as depicted in Fig. \ref{fig:normal_contact1}. It is important to note that this approximation of complementarity is not perfectly accurate in the vicinity of the critical state. In situations where $c$ is chosen to be extremely small, significant errors may occur when $x$ approaches zero. In practical simulation computations, it is essential to meticulously select the parameter $c$ to strike a balance between accuracy and the numerical robustness of the calculations.
\subsubsection{Trigonometric function}
The second smooth function we proposed is based on trigonometric function as follows: 
\begin{equation}\label{aaa}
	D(x)=\left\{\begin{aligned}
		&	0\ , \ x<0\\
		&	\frac{1-\cos(wx)}{2} \ , 0\leq\ x<\frac{\pi}{w}\\
		&   1\ , \ x\geq\frac{\pi}{w}
	\end{aligned}\right.
\end{equation}
where $w$ is a positive parameter.

In contrast to the sigmoid function, the system described by (\ref{aaa}) rigorously fulfills the complementarity condition. This is attributed to their positive definiteness in the positive half-axis of $x$, while being identically zero in the negative half-axis of $x$.

\begin{figure}[h]
	\centering
	\includegraphics[width=0.49\textwidth]{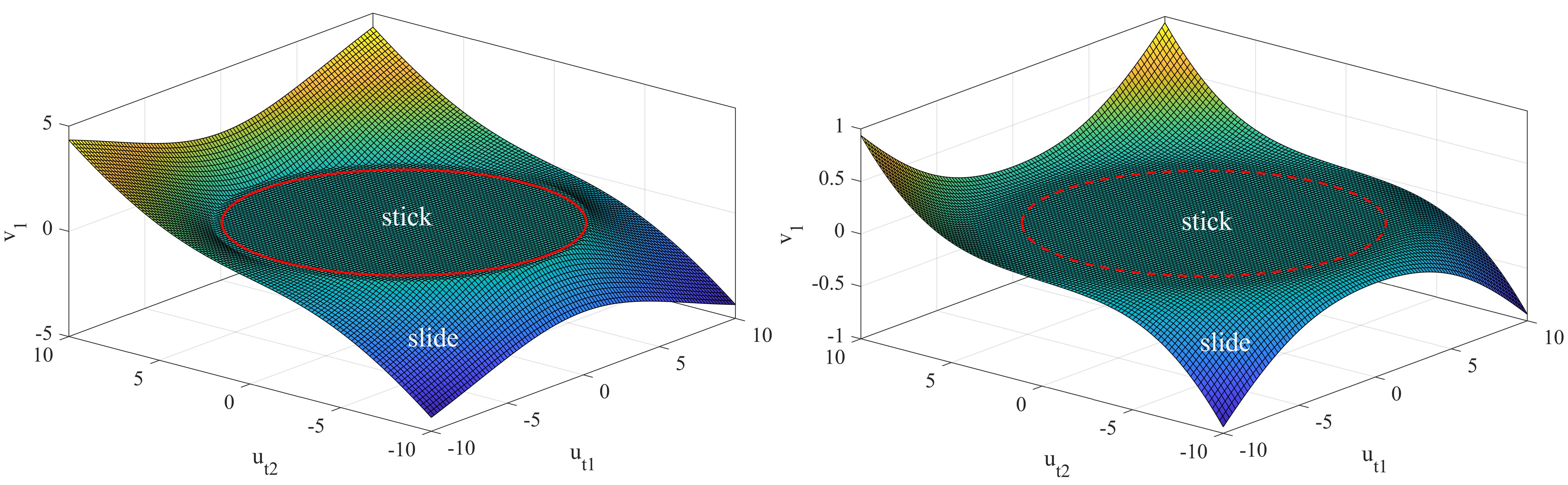}
	\caption{The manifold of tangent velocity with respect of slack variable. The figure illustrates a comparison between the non-smooth case utilizing the Heaviside function (on the left) and the smooth case employing a trigonometric function (on the right).}
	\label{fig:tangent_velocity}
\end{figure}

By utilizing the aforementioned function as a substitute for the impact function, we obtain the smoothed contact constraints, as depicted in Fig. \ref{fig:tangent_velocity}. At the critical state between sticking and sliding, a sudden change in gradient can be observed in the left image of Fig. \ref{fig:tangent_velocity} which is non smooth, whereas the gradient in the right image of Fig. \ref{fig:tangent_velocity} exhibits a continuous transition using trigonometric function.
\section{Cosserat dynamics with contact}\label{sec::dynamics}
After addressing all types of contact constraints, this section will primarily focus on the kinematics and dynamics of soft robots under contact constraints.
\subsection{Continuous kinematics}
It was derived in \cite{8500341} that the continuous kinematics model is represented by the following two differential equations: 
\begin{equation}\label{geom2}
	\boldsymbol{g}^{\prime}=\boldsymbol{g} \widehat{\boldsymbol{\xi}}
\end{equation}
\begin{equation}\label{velo}
	\boldsymbol{\eta}^{\prime}=\dot{\boldsymbol{\xi}}-\operatorname{ad}_{\boldsymbol{\xi}} \boldsymbol{\eta}
\end{equation}
The significance of the above kinematic equations lies in the fact that the complete velocity field of a soft slender robot can be determined from the strain field and the rate of change of the strain field over time. These equations establish a fundamental relationship between the deformation of the soft slender robot and its corresponding motion. 
\subsection{Strong form of dynamics}
\subsubsection{PDE}
The corresponding dynamic model {written in the} body frame was derived in \cite{black2017parallel} and \cite{boyer2020dynamics} as the following partial differential equations (PDEs):
\begin{equation}\label{pde}
	\begin{aligned}
	\boldsymbol{\mathcal{M}}\dot{\boldsymbol{\eta}}-ad_{\boldsymbol{\eta}}^{\top}\boldsymbol{\mathcal{M}}\boldsymbol{\eta}&=\boldsymbol{\Lambda}_{i}^{\prime}-\operatorname{ad}_{\boldsymbol{\xi}}^{\top} \boldsymbol{\Lambda}_{i}+{\boldsymbol{\Lambda}}_e\\&+\mathrm{Ad}^*_{\boldsymbol{g}_{bc}}{\boldsymbol{\Lambda}}_c+{{\boldsymbol{\Lambda}}_{f}+\mathrm{Ad}^*_{\boldsymbol{g}_{bc}}\mathbf{B}^\top{\boldsymbol{\Lambda}}_{a}}
	\end{aligned}
\end{equation}
satisfying the following boundary conditions:
\begin{equation}\label{key}
	\begin{aligned}
		\boldsymbol{\Lambda}_{i}(0,t) = -{\boldsymbol{\Lambda}}_{0}(t)\ , \ \mathrm{or} \ \boldsymbol{g}(0,t)=\boldsymbol{g}_0(t)\\
		\boldsymbol{\Lambda}_{i}(1,t) = {\boldsymbol{\Lambda}}_{1}(t)\ , \ \mathrm{or} \ \boldsymbol{g}(1,t)=\boldsymbol{g}_1(t)
	\end{aligned}
\end{equation}
where $\boldsymbol{\mathcal{M}}\in \mathbb{R}^{6\times6}$ is the tensor of mass linear density along central axis. $\boldsymbol{\Lambda}_{i}\in \mathbb{R}^{6}$ is the elastic internal wrench. ${\boldsymbol{\Lambda}}_{c}\in \mathbb{R}^{6}$ is the contact force produced by collision. {${\boldsymbol{\Lambda}}_{f}\in \mathbb{R}^{6}$ is the contact force produced by fixed constraints and ${\boldsymbol{\Lambda}}_{a}\in \mathbb{R}^{3}$ is the contact force produced by articulated constraints.  ${\boldsymbol{\Lambda}}_e\in \mathbb{R}^{6}$ are the distributed external wrench applied on the soft slender robot.} One contribution of ${\boldsymbol{\Lambda}}_e$ comes from the gravity which can be given by
${\boldsymbol{\Lambda}}_e=\boldsymbol{\mathcal{M}}\mathrm{Ad}_{\boldsymbol{g}}^{-1}\boldsymbol{\mathcal{G}}$, where the inverse of adjoint representation of the Lie group is
used to transform twists from the global frame to the body frame. $\boldsymbol{\mathcal{G}}$ is the gravity acceleration twist w.r.t. the global frame $\boldsymbol{\mathcal{G}}=[\boldsymbol{0}_{1\times 5} \ -9.81]^\top$.
\subsubsection{Constraints}
If all contact constraints introduced in section \ref{sec::cont_constrain} are considered, the dynamics must also satisfy the following constraint conditions:
\begin{equation}\label{44}
	{\boldsymbol{G}}_{f}=\boldsymbol{0}
\end{equation}
\begin{equation}\label{66}
	{\boldsymbol{G}}_{a}=\boldsymbol{0}
\end{equation}
\begin{equation}\label{lc}
	{\boldsymbol{\Lambda}}_c-(\mathbf{B}_n^\top D\mathbf{C}_n+\mathbf{B}_t^\top W\mathbf{C}_t)\boldsymbol{u}=\boldsymbol{0}
\end{equation}
\begin{equation}\label{cont1}
	\mathbf{P}\boldsymbol{g}_{cd}\mathbf{A}+D^*\mathbf{C}_n\boldsymbol{u}=0
\end{equation}
\begin{equation}\label{cont2}
	\mathrm{Ad}_{\boldsymbol{g}_{cb}}\boldsymbol{\eta}-\mathrm{Ad}_{\boldsymbol{g}_{cd}}\boldsymbol{\eta}_d-W^*\mathbf{C}_t\boldsymbol{u}=\boldsymbol{0}
\end{equation}
(\ref{44}) and (\ref{66}) contains all bilateral constraints and (\ref{lc})-(\ref{cont2}) denote the reformulation of the unilateral constraint based on the slack variable. (\ref{lc}) represents the constraint of contact force. (\ref{cont1}) and (\ref{cont2}) describe the constraint on normal contact gap and the constraint on tangential sliding respectively. 
The unknown variables in the partial differential-algebraic system formed by the dynamics and constraints discussed above (\ref{geom2})-(\ref{cont2}) encompass the initial pose tensor $\boldsymbol{g}_0(t)$ at the starting end, the strain field $\boldsymbol{\xi}(s,t)$, the contact force $\boldsymbol{\Lambda}_f(s,t)$, $\boldsymbol{\Lambda}_a(s,t)$ and the contact field represented by slack variables $\boldsymbol{u}(s,t)$. In order to facilitate the solution of this partial differential-algebraic system, we aim to derive its corresponding weak form. By introducing suitable trial functions {$\boldsymbol{\Phi}_{\boldsymbol{\xi}}(s)$ and $\boldsymbol{\Phi}_{u}(s)$}, the weak form of (\ref{pde}) can be expressed as the integration through the backbone of soft slender robot:
\begin{equation}\label{eq:dynamic_weak1}
	\begin{aligned}	
		\int_{0}^{1}&\boldsymbol{\Phi}_{\boldsymbol{\xi}}^\top(\boldsymbol{\mathcal{M}}\dot{\boldsymbol{\eta}}-ad_{\boldsymbol{\eta}}^{\top}\boldsymbol{\mathcal{M}}\boldsymbol{\eta}-\boldsymbol{\Lambda}_{i}^{\prime}+\operatorname{ad}_{\boldsymbol{\xi}}^{\top} \boldsymbol{\Lambda}_{i}-{\boldsymbol{\Lambda}}_e\\
		&-\mathrm{Ad}^*_{\boldsymbol{g}_{bc}}{\boldsymbol{\Lambda}}_{c}-{{\boldsymbol{\Lambda}}_{f}-\mathrm{Ad}^*_{\boldsymbol{g}_{bc}}\mathbf{B}^\top{\boldsymbol{\Lambda}}_{a}})\mathrm{ds}=0
	\end{aligned}
\end{equation}
and the weak form of constraints (\ref{cont1}) and (\ref{cont2}) are given by: 
\begin{equation}\label{w11}
	\int_{0}^{1}\boldsymbol{\Phi}_{\boldsymbol{u}}^\top
	\begin{bmatrix}
		\mathbf{P}\boldsymbol{g}_{cd}\mathbf{A}-D^*\mathbf{C}_n\boldsymbol{u}\\
		\mathrm{Ad}_{\boldsymbol{g}_{cb}}\boldsymbol{\eta}-\mathrm{Ad}_{\boldsymbol{g}_{cd}}\boldsymbol{\eta}_d-W^*\mathbf{C}_t\boldsymbol{u}
	\end{bmatrix}\mathrm{ds}=0
\end{equation}
{As for the bilateral constrain (\ref{44}) and (\ref{66}), their weak forms retain the same representation as the strong form since they are normally discretely located on the soft slender robot.} 
Applying discretization techniques, the above integral equations of weak forms are further approximated. This discretization process involves dividing the spatial space of robot's backbone into discrete nodes, thereby transforming the weak form into a more tractable differential-algebraic system. 
\section{Spatial discritization of strain and contact}\label{sec::Discritization}
\subsection{Spatial interpolation of soft slender robot}
In the subsequent discussion, we propose a redefinition of the robot's configuration space by parameterizing $\boldsymbol{g}_0(t)$ and $\boldsymbol{\xi}(s,t)$. $\boldsymbol{g}_0$ is composed of a rotation matrix $\boldsymbol{R}_0$ and a position vector $\boldsymbol{p}_0$. In our approach, we use the exponential map to define the rotation matrix $\boldsymbol{R}_0$ by a vector $\boldsymbol{\phi}\in \mathbb{R}^3$:
\begin{equation}\label{g0}
	\boldsymbol{R}_0(t)=\exp\tilde{\boldsymbol{\phi}}(t)\ , \ \boldsymbol{\phi}(t)\in \mathbb{R}^3
\end{equation}
Thus the configuration matrix $\boldsymbol{g}_0(t)$ is parameterized by the vector $\boldsymbol{\phi}(t)$ and $\boldsymbol{p}_0(t)$. We use vector $\boldsymbol{\alpha}(t)\in\mathbb{R}^6$ to denote their combination, i.e., $\boldsymbol{\alpha}(t)=[\boldsymbol{\phi}^\top(t) \ , \  \boldsymbol{p}_0^\top(t)]^\top$. 

For parameterizing the strain field, we adopt the piecewise linear strain (PLS) assumption to interpolate the strain field. The details of PLS method can be found in our previous modeling work \cite{10027557}. Dividing the soft slender robot into $p$ sections along the $s$ direction in the form of $[0,s_1]$, $[s_1,s_2]$, $\dots$, $[s_{p-1},1]$, the entire strain field is reformulated by a set of linear interpolation basis functions $\boldsymbol{\Phi}(s)\in\mathbb{R}^{6\times6(p+1)}$ and a discrete set of strain vectors $\boldsymbol{\theta}\in\mathbb{R}^{6(p+1)}$:
\begin{equation}\label{key}
	\boldsymbol{\xi}(s,t)=\boldsymbol{\xi}_0+\boldsymbol{\Phi}(s)\boldsymbol{\theta}(t)
\end{equation}
with $\boldsymbol{\theta}(t)=[\boldsymbol{\xi}^\top(0,t),\boldsymbol{\xi}^\top(s_1,t),\dots,\boldsymbol{\xi}^\top(1,t)]^\top$.
By employing the aforementioned parameterization, the entire configuration space of soft slender robot is now reconstructed in the following form:
\begin{equation}\label{key}
	\mathcal{U}_d=SE(3)\times\mathbb{R}^{6(p+1)}
\end{equation}
where $SE(3)$ stands for the configuration $\boldsymbol{g}_0$ parameterized via (\ref{g0}).
Therefore, by numerically integrating (\ref{geom2}) in $SE(3)$, the geometric model of the slender robot can be expressed by the following equation:\\
$\forall s\in [s_{k-1},s_k]$,
\begin{equation}\label{gg}
	\boldsymbol{g}(s,t)=\boldsymbol{g}_0(t)\left(\prod_{i=1}^{k-1}\mathrm{exp}\widehat{\boldsymbol{\xi}}(s_i,t)\mathrm{\Delta s})\right)\mathrm{exp}\widehat{\boldsymbol{\xi}}(s-s_k,t)
\end{equation}
where $\Delta s$ is the integrate segment along length of slender robot.
\subsection{Discrete kinematics}
As a common way in robotics community, Jacobian matrix is defined to map the generalized coordinates $\boldsymbol{q}(t)=[\boldsymbol{\alpha}^\top(t)  ,  \boldsymbol{\theta}^\top(t)]^\top\in \mathbb{R}^{6(p+2)}$ to velocity field $\boldsymbol{\eta}(s,t)\in \mathbb{R}^6$ via the following form:
\begin{equation}\label{99}
	\boldsymbol{\eta}(s,t)=\underbrace{\begin{bmatrix}
			\boldsymbol{J}_{\boldsymbol{\alpha}}(s,t) & \boldsymbol{J}_{\boldsymbol{\theta}}(s,t)
	\end{bmatrix}}_{\boldsymbol{J}(s,t)\in \mathbb{R}^{6\times 6(p+2)}}\dot{\boldsymbol{q}}(t)
\end{equation}
where $\boldsymbol{J}_{\boldsymbol{\alpha}}(s,t)\in \mathbb{R}^{6\times6}$ and $\boldsymbol{J}_{\boldsymbol{\theta}}(s,t)\in \mathbb{R}^{6\times6(p+1)}$ are two Jacobian matrices respectively related with $\boldsymbol{\alpha}(t)$ and $\boldsymbol{\theta}(t)$. Then, by substituting (\ref{99}) into (\ref{velo}), one can get:
$$\boldsymbol{J}_{\boldsymbol{\alpha}}^\prime\dot{\boldsymbol{\alpha}}+\boldsymbol{J}_{\boldsymbol{\theta}}^\prime\dot{\boldsymbol{\theta}}=\boldsymbol{\Phi}\dot{\boldsymbol{\theta}}-\operatorname{ad}_{\boldsymbol{\xi}} (\boldsymbol{J}_{\boldsymbol{\alpha}}\dot{\boldsymbol{\alpha}}+\boldsymbol{J}_{\boldsymbol{\theta}}\dot{\boldsymbol{\theta}})$$
Note that the above equation holds for any $\dot{\boldsymbol{\alpha}}$ and $\dot{\boldsymbol{\theta}}$, thus the following two equations stand:
\begin{equation}\label{1}
	\boldsymbol{J}_{\boldsymbol{\theta}}^\prime=-\operatorname{ad}_{\boldsymbol{\xi}}\boldsymbol{J}_{\boldsymbol{\theta}}+\boldsymbol{\Phi}
\end{equation}
\begin{equation}\label{2}
	\boldsymbol{J}_{\boldsymbol{\alpha}}^\prime=-\operatorname{ad}_{\boldsymbol{\xi}}\boldsymbol{J}_{\boldsymbol{\alpha}}
\end{equation}
Taking derivative of the two equations above with respect to time one can get:
\begin{equation}\label{3}		          \dot{\boldsymbol{J}}_{\boldsymbol{\theta}}^\prime=-\operatorname{ad}_{\dot{\boldsymbol{\xi}}}\boldsymbol{J}_{\boldsymbol{\theta}}-\operatorname{ad}_{\boldsymbol{\xi}}\dot{\boldsymbol{J}}_{\boldsymbol{\theta}}
\end{equation}
\begin{equation}\label{4}		          \dot{\boldsymbol{J}}_{\boldsymbol{\alpha}}^\prime=-\operatorname{ad}_{\dot{\boldsymbol{\xi}}}\boldsymbol{J}_{\boldsymbol{\alpha}}-\operatorname{ad}_{\boldsymbol{\xi}}\dot{\boldsymbol{J}}_{\boldsymbol{\alpha}}
\end{equation}
Then, the Jacobian matrices and their time derivatives can be computed by numerical integration along arc space $s\in[0,1]$ through (\ref{1})-(\ref{4}) respectively starting from the initial value of $\boldsymbol{J}_{\boldsymbol{\alpha}}(0,t)$ and $\boldsymbol{J}_{\boldsymbol{\theta}}(0,t)$. It is obvious that for $s=0$ the velocity $\boldsymbol{\eta}(0,t)$ only depends on the base parameter $\boldsymbol{\alpha}(t)$, i.e., $\boldsymbol{\eta}(0,t)=\boldsymbol{J}_{\boldsymbol{\alpha}}(0,t)\dot{\boldsymbol{\alpha}}$, thus we can directly deduce $\boldsymbol{J}_{\boldsymbol{\theta}}(0,t)=\boldsymbol{0}$ and $\dot{\boldsymbol{J}}_{\boldsymbol{\theta}}(0,t)=\boldsymbol{0}$.
For the Jacobian with respect to $\boldsymbol{\alpha}(t)$ at the initial position, i.e., $\boldsymbol{J}_{\boldsymbol{\alpha}}(0,t)$, it can be derived from the derivation of exponential map. 

For the exponential map of rotation matrix, the following differential relationship stands:
\begin{equation}\label{tt}
	\dot{\boldsymbol{R}}_0(\boldsymbol{\phi})\boldsymbol{R}_0^\top(\boldsymbol{\phi})=\widetilde{\Big(\boldsymbol{J}_l(\boldsymbol{\phi})\dot{\boldsymbol{\phi}}(t)\Big)}
\end{equation}
where $(\tilde{\cdot})$ denotes the mapping from $\mathbb{R}^3$ to $SO(3)$.  $\boldsymbol{J}_l$ is the left Jacobian of group $SO(3)$, with the definition as below:
$$\boldsymbol{J}_l(\boldsymbol{\phi}) = \mathbf{I}+\frac{1-\cos\phi}{\phi^2}\tilde{\boldsymbol{\phi}}+\frac{\phi-\sin\phi}{\phi^3}\tilde{\boldsymbol{\phi}}^2$$
Using (\ref{tt}), the angular velocity of initial position with respect to body frame can be deduced: 
$$\tilde{\boldsymbol{w}}(0,t)=\boldsymbol{R}_0^\top\dot{\boldsymbol{R}}_0=\boldsymbol{R}_0^\top(\widetilde{\boldsymbol{J}_l\dot{\boldsymbol{\phi}}})\boldsymbol{R}_0$$
Notice that for any vector $\boldsymbol{x}\in \mathbb{R}^3$ and matrix $\boldsymbol{R}\in SO(3)$, one holds:
$\boldsymbol{R}^\top\tilde{\boldsymbol{x}}\boldsymbol{R}=(\widetilde{\boldsymbol{R}^\top\boldsymbol{x}})$.
Thus one can get
\begin{equation}\label{key}
	\boldsymbol{w}(0,t)=\boldsymbol{R}_0^\top(\boldsymbol{\phi})\boldsymbol{J}_l(\boldsymbol{\phi})\dot{\boldsymbol{\phi}}
\end{equation}
The linear velocity of initial position with respect to body frame is 
$$\boldsymbol{v}(0,t)=\boldsymbol{R}_0^\top(\boldsymbol{\phi})\dot{\boldsymbol{p}}_0$$
We can finally deduce the map from time derivation of vector $\boldsymbol{\alpha}$ to velocity twist of initial position with respect to body frame:
\begin{equation}\label{key}
	\boldsymbol{\eta}(0,t)=\begin{bmatrix}
		\boldsymbol{R}_0^\top(\boldsymbol{\phi})\boldsymbol{J}_l(\boldsymbol{\phi})&\boldsymbol{0}\\
		\boldsymbol{0}&\boldsymbol{R}_0^\top(\boldsymbol{\phi})
	\end{bmatrix}\dot{\boldsymbol{\alpha}}(t)
\end{equation}
where the matrix mapping $\dot{\boldsymbol{\alpha}}(t)$ to $\boldsymbol{\eta}(0,t)$ stands for the Jacobian $\boldsymbol{J}_{\boldsymbol{\alpha}}(0,t)$. 
\subsection{Spatial interpolation of contact field}
It is clear that the distribution of contact forces depends on the arc length parameter $s$, thus similar to the interpolation of the strain field, we employ the concept of piecewise linear interpolation to approximate the distribution of contact forces through interpolating the corresponding slack variable along the length of the center line. By dividing the soft slender robot into $m$ sections along the $s$ direction, denoted as $[0,s_1]$, $[s_1,s_2]$, $\dots$, $[s_{m-1},1]$, the entire contact field ${\boldsymbol{\Lambda}}_c(s)$ can be expressed as a linear combination of interpolation basis functions $\boldsymbol{\Psi}(s)$ and a discrete set of slack variable $\boldsymbol{\lambda}_{c}$:
\begin{equation}\label{key}
	{\boldsymbol{\Lambda}}_c(s)=(\mathbf{B}_n^\top D\mathbf{C}_n+\mathbf{B}_t^\top W\mathbf{C}_t)\boldsymbol{\Psi}(s){\boldsymbol{\lambda}}_c
\end{equation}
where
${\boldsymbol{\lambda}}_{c}=[{u}_{0}\ {u}_{1}\ \dots\ {u}_{m}]^\top\in\mathbb{R}^{m+1}$ contains the slack variables of each node along centerline. 
The interpolation basis functions $\boldsymbol{\Psi}(s)$ capture the linear variation of contact forces within each section of the soft slender robot, which are defined as follows:
$$\boldsymbol{\Psi}(s)=\begin{bmatrix}
	{\Psi}_{0}\mathbf{I}_{3\times 3}&{\Psi}_{1}\mathbf{I}_{3\times 3}&\dots&{\Psi}_{m}\mathbf{I}_{3\times 3}
\end{bmatrix}\in\mathbb{R}^{3\times 3(m+1)}$$
where for $i\in\{0,1,\dots,m\}$
$$\Psi_{i}(s)=\left\{\begin{aligned}
	ms-i& \ \ \mathrm{for} \ s\in[\frac{i-1}{m},\frac{i}{m})\\
	i+1-ms& \ \ \mathrm{for} \ s\in[\frac{i}{m},\frac{i+1}{m}]\\
	0& \ \ \mathrm{for} \ s\notin[\frac{i-1}{m},\frac{i+1}{m}]
\end{aligned}
\right.$$
\begin{figure}[H]
	\centering
	\includegraphics[width=0.40\textwidth]{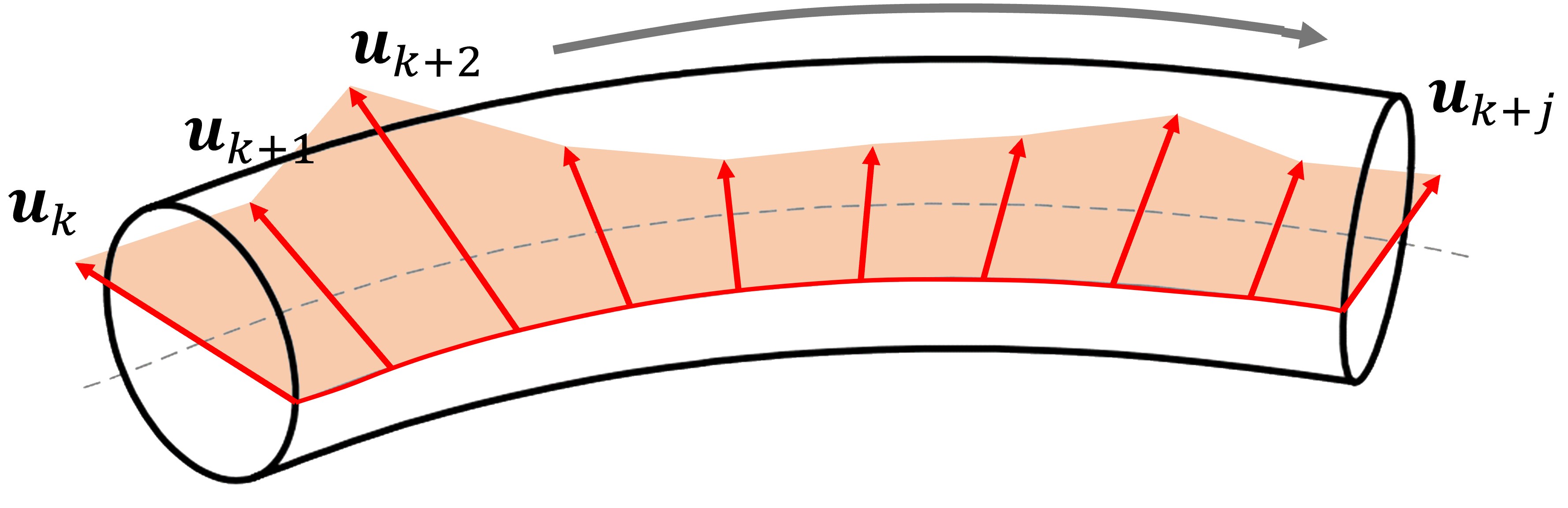}
	\caption{Linear interpolation of contact load.}
	\label{fig:linearcontact}
\end{figure}
{As for the contact force produced by fixed constraints and articulated constraints, supposing that they are discretely located on the soft slender robot with quantities of $m_f$ and $m_a$ respectively, we use two vectors to cover all of them:
$${\boldsymbol{\lambda}}_{f}=[\boldsymbol{\Lambda}_{f,1}^\top\ \boldsymbol{\Lambda}_{f,2}^\top\ \dots\ \boldsymbol{\Lambda}_{f,m_f}^\top]^\top\in\mathbb{R}^{6m_f}$$
$${\boldsymbol{\lambda}}_{a}=[\boldsymbol{\Lambda}_{a,1}^\top\ \boldsymbol{\Lambda}_{a,2}^\top\ \dots\ \boldsymbol{\Lambda}_{a,m_a}^\top]^\top\in\mathbb{R}^{6m_a}$$}
\subsection{Weak form of dynamics}
For $s\in[0,1]$, by introducing the test function $\boldsymbol{\Phi}_{\boldsymbol{\xi}}(s)$ as virtual displacement $\boldsymbol{\delta}(s)=\boldsymbol{J}\delta\boldsymbol{q}$, the weak form (\ref{eq:dynamic_weak1}) is given by:   
\begin{equation}\label{eq:dynamic_weak}
	\begin{aligned}	
		\delta\boldsymbol{q}^\top\int_{0}^{1}\boldsymbol{J}^\top&(\boldsymbol{\mathcal{M}}\dot{\boldsymbol{\eta}}-ad_{\boldsymbol{\eta}}^{\top}\boldsymbol{\mathcal{M}}\boldsymbol{\eta}-\boldsymbol{\Lambda}_{i}^{\prime}+\operatorname{ad}_{\boldsymbol{\xi}}^{\top} \boldsymbol{\Lambda}_{i}-{\boldsymbol{\Lambda}}_e\\&-\mathrm{Ad}^*_{\boldsymbol{g}_{bc}}{\boldsymbol{\Lambda}}_c-{{\boldsymbol{\Lambda}}_{f}-\mathrm{Ad}^*_{\boldsymbol{g}_{bc}}\mathbf{B}^\top{\boldsymbol{\Lambda}}_{a}})\mathrm{ds}=0
	\end{aligned}
\end{equation}
By introducing the test function $\boldsymbol{\Phi}_{\boldsymbol{u}}(s)$ as $\delta \boldsymbol{u}=\boldsymbol{\Psi}\delta \boldsymbol{\lambda}$ for contact constrain (\ref{cont1}) and (\ref{cont2}) respectively, their weak form (\ref{w11}) is deduced as follows:
\begin{equation}\label{w22}
	\delta\boldsymbol{\lambda}_c^\top\int_{0}^{1}\boldsymbol{\Psi}^\top
	\begin{bmatrix}
		\mathbf{P}\boldsymbol{g}_{cd}\mathbf{A}+D^*\mathbf{C}_n\boldsymbol{\Psi}{\boldsymbol{\lambda}}_c\\
		\mathrm{Ad}_{\boldsymbol{g}_{cb}}\boldsymbol{\eta}-\mathrm{Ad}_{\boldsymbol{g}_{cd}}\boldsymbol{\eta}_d-W^*\mathbf{C}_t\boldsymbol{\Psi}{\boldsymbol{\lambda}}_c
	\end{bmatrix}\mathrm{ds}=0
\end{equation}
\subsection{Discrete dynamics equation}
By reformulating the contact force using $\boldsymbol{\lambda}_n$, $\boldsymbol{\lambda}_t$, $\boldsymbol{\lambda}_{fc}$ and $\boldsymbol{\lambda}_{ac}$, along with the aforementioned definitions, we have successfully transformed the dynamics from the NCP form into a system of DAEs (Differential-Algebraic Equations) solely composed of equality constraints:
\begin{equation}\label{dynamic}
	\begin{aligned}
		\boldsymbol{M}\ddot{\boldsymbol{q}}+\boldsymbol{C}\dot{\boldsymbol{q}}+\boldsymbol{K}\boldsymbol{q}=\boldsymbol{P}+\boldsymbol{H}_c{\boldsymbol{\lambda}}_c+{\boldsymbol{H}_{f}{\boldsymbol{\lambda}}_{f}+\boldsymbol{H}_{a}{\boldsymbol{\lambda}}_{a}}
	\end{aligned}
\end{equation}
\begin{equation}\label{key}
	\boldsymbol{K}_{BC}\boldsymbol{q}-\boldsymbol{\Lambda}_{BC}=\boldsymbol{0}
\end{equation}
\begin{equation}\label{key1}		                         \boldsymbol{G}_c-\boldsymbol{E}_c\boldsymbol{\lambda}_c=\boldsymbol{0}
\end{equation}
\begin{equation}\label{gf}		                         \boldsymbol{G}_{f}=\boldsymbol{0}
\end{equation}
\begin{equation}\label{key3}		                         \boldsymbol{G}_{a}=\boldsymbol{0}
\end{equation}
{For ease of description, we define the following operators $\mathrm{Vec}$ to assemble discrete matrices together:
$$\overset{N}{\underset{i=1}{\mathrm{Vec}}}(\boldsymbol{X}_i)=[\boldsymbol{X}_1^\top \ \boldsymbol{X}_2^\top \ \dots \ \boldsymbol{X}_N^\top]^\top$$}
All the matrices of (\ref{dynamic})-(\ref{key3}) are given by the following definitions:
{\begin{itemize}
	\item[$\bullet$]{$\boldsymbol{M}(\boldsymbol{q})=\int_0^1\boldsymbol{J}^\top\boldsymbol{\mathcal{M}}\boldsymbol{J}ds\in \mathbb{R}^{6(n+1)\times 6(n+1)}$}, the mass matrix;
	\item[$\bullet$]{$\boldsymbol{C}(\boldsymbol{q},\dot{\boldsymbol{q}})=\int_0^1\boldsymbol{J}^\top(\boldsymbol{\mathcal{M}}\dot{\boldsymbol{J}}- ad_{\boldsymbol{J}\dot{\boldsymbol{q}}}^{\top}\boldsymbol{\mathcal{M}}\boldsymbol{J})ds\in \mathbb{R}^{6(n+1)\times 6(n+1)}$}, the Coriolis matrix;
	\item[$\bullet$]{$\boldsymbol{K}=\int_0^1\boldsymbol{\Phi}^\top\mathcal{K}\boldsymbol{\Phi}ds\in \mathbb{R}^{6(n+1)\times 6(n+1)}$}, the stiffness matrix;

{ 	   
	\item[$\bullet$]{$\boldsymbol{H}_c(\boldsymbol{q})=\int_{0}^{1}\boldsymbol{J}^\top\mathrm{Ad}^*_{\boldsymbol{g}_{bc}}(\mathbf{B}_n^\top D\mathbf{C}_n+\mathbf{B}_t^\top W\mathbf{C}_t)\boldsymbol{\Psi}\mathrm{ds}\in \mathbb{R}^{6(n+1)\times 3(m+1)}$}, the collision contact force matrix;}
	{\item[$\bullet$]{$\boldsymbol{H}_{f}(\boldsymbol{q})=\overset{m_f}{\underset{i=1}{\mathrm{Vec}}}(\boldsymbol{J}_i)\in \mathbb{R}^{6(n+1)\times 6m_f}$}, the force matrix of fixed constraints;
	\item[$\bullet$]{$\boldsymbol{H}_{a}(\boldsymbol{q})=\overset{m_a}{\underset{i=1}{\mathrm{Vec}}}(\boldsymbol{J}_i^\top\mathrm{Ad}^*_{\boldsymbol{g}_{bc,i}}\mathbf{B}_i^\top)\in \mathbb{R}^{6(n+1)\times 3m_a}$}, the force matrix of articulated constraints;}
	\item[$\bullet$]{$\boldsymbol{P}(\boldsymbol{q})=\int_{0}^{1}\boldsymbol{J}^\top{\boldsymbol{\Lambda}}_e\mathrm{ds}+\int_{0}^{1}\boldsymbol{J}^\top\boldsymbol{\mathcal{M}}\mathrm{Ad}_{\boldsymbol{g}}^{-1}\mathrm{ds}\boldsymbol{\mathcal{G}}\in \mathbb{R}^{6(n+1)}$}, the contribution of concentrated external load and gravity.
	\item[$\bullet$]{$\boldsymbol{K}_{BC}=\mathrm{diag}\big(\mathcal{K}(0)\boldsymbol{\Phi}(0),\mathcal{K}(1)\boldsymbol{\Phi}(1)\big)\in \mathbb{R}^{12\times12}$}, the stiffness matrix of boundary condition;
	\item[$\bullet$]{$\boldsymbol{\Lambda}_{BC}=[-\boldsymbol{\Lambda}_0^\top \ \boldsymbol{\Lambda}_1^\top]^\top\in \mathbb{R}^{12}$}, the external force of boundary;

{ 
	\item[$\bullet$]{$\boldsymbol{G}_c(\boldsymbol{q},\dot{\boldsymbol{q}})=\int_{0}^{1}\boldsymbol{\Psi}^\top
	\begin{bmatrix}
		\mathbf{P}\boldsymbol{g}_{cd}\mathbf{A}\\
		\mathrm{Ad}_{\boldsymbol{g}_{cb}}\boldsymbol{\eta}-\mathrm{Ad}_{\boldsymbol{g}_{cd}}\boldsymbol{\eta}_d
	\end{bmatrix}\mathrm{ds}\in \mathbb{R}^{3(m+1)}$}, the tangent contact velocity matrix;}
	{\item[$\bullet$]{$\boldsymbol{G}_{f}(\boldsymbol{q})=\overset{N}{\underset{i=1}{\mathrm{Vec}}}\left(\log(\boldsymbol{g}_{fc,i}^{-1}\boldsymbol{g}_i)\right)^\vee\in \mathbb{R}^{6m_f}$}, the fixed constraint vector;
	\item[$\bullet$]{$\boldsymbol{G}_{a}(\boldsymbol{q})=\overset{N}{\underset{i=1}{\mathrm{Vec}}}(\boldsymbol{p}_{a,i}-\mathbf{U}\boldsymbol{g}_{c,i}\mathbf{A})\in \mathbb{R}^{6m_a}$}, the articulated constraint vector;}

	{\item[$\bullet$]{$\boldsymbol{E}_c(\boldsymbol{\lambda}_c)=\int_{0}^{1}\boldsymbol{\Psi}^\top
		\begin{bmatrix}
			D^*\mathbf{C}_n\\
			-W^*\mathbf{C}_t
		\end{bmatrix}\boldsymbol{\Psi}\mathrm{ds}\in \mathbb{R}^{3(m+1)\times 3(m+1)}$}, the collision contact constraint matrix;}
\end{itemize}}
\section{Time discretization}\label{sec::implicit}
Time-stepping is a prevalent technique employed for the time discretization of dynamic systems. It has gained significant popularity in the field of robotics for simulation and control purposes.
Considering a time interval $\begin{bmatrix}
	t_{k-1},t_k
\end{bmatrix}$ and denoting $h=t_k-t_{k-1}$ as the time step,  
for the explicit representation, the discritization of this time interval is given by: $$\boldsymbol{q}_{k}=\boldsymbol{q}_{k-1}+h\dot{\boldsymbol{q}}_{k-1}\ , \ \dot{\boldsymbol{q}}_{k}=\dot{\boldsymbol{q}}_{k-1}+h\ddot{\boldsymbol{q}}_{k-1}$$
For the implicit representation, the discritization of this time interval is given by:
$$\boldsymbol{q}_{k}=\boldsymbol{q}_{k-1}+h\dot{\boldsymbol{q}}_{k}\ , \ \dot{\boldsymbol{q}}_{k}=\dot{\boldsymbol{q}}_{k-1}+h\ddot{\boldsymbol{q}}_{k}$$
In our work, we use the implicit representation to ensure convergence of the solution.
By taking the implicit equations to (\ref{dynamic}) and using the abbreviation $\widehat{\boldsymbol{M}}=\boldsymbol{M}+h\boldsymbol{C}$, one can get:
$$\boldsymbol{\mathscr{L}}(\dot{\boldsymbol{q}}_k,{\boldsymbol{q}}_k)=\widehat{\boldsymbol{M}}\dot{\boldsymbol{q}}_{k}-\boldsymbol{M}\dot{\boldsymbol{q}}_{k-1}-h\boldsymbol{K}\boldsymbol{q}_{k}-h\boldsymbol{P}$$
Denoting $$\boldsymbol{\mathscr{F}}(\boldsymbol{q}_k,\boldsymbol{\lambda}_c,\boldsymbol{\lambda}_f,\boldsymbol{\lambda}_a)=\boldsymbol{P}+\boldsymbol{H}_c{\boldsymbol{\lambda}}_c+{\boldsymbol{H}_{f}{\boldsymbol{\lambda}}_{f}+\boldsymbol{H}_{a}{\boldsymbol{\lambda}}_{a}}$$
the implicit time discretization of (\ref{dynamic})-(\ref{key3}) is replaced by the following nonlinear algebraic equations:
\begin{subnumcases}{\label{governing2}}
	\boldsymbol{\mathscr{L}}(\dot{\boldsymbol{q}}_k,{\boldsymbol{q}}_k)-h\boldsymbol{\mathscr{F}}(\boldsymbol{q}_k,\boldsymbol{\lambda}_c,\boldsymbol{\lambda}_f,\boldsymbol{\lambda}_a)=\mathbf{0}\label{dynnn1}\\
	\boldsymbol{q}_k-h\dot{\boldsymbol{q}}_k-\boldsymbol{q}_{k-1}=\mathbf{0}\label{im1p}
	\\		\boldsymbol{K}_{BC}\boldsymbol{q}_k-\boldsymbol{\Lambda}_{BC}=\boldsymbol{0}
	\\
	\boldsymbol{G}_c(\boldsymbol{q}_k,\dot{\boldsymbol{q}_k})-\boldsymbol{E}_c\boldsymbol{\lambda}_c=\boldsymbol{0}\label{ncp111}
	\\
	\boldsymbol{G}_{f}(\boldsymbol{q}_k)=\mathbf{0}\\
	\boldsymbol{G}_{a}(\boldsymbol{q}_k)=\mathbf{0}
\end{subnumcases}
The unknown variables in the algebraic equations (\ref{governing2}) consist of $\boldsymbol{q}_k$, $\dot{\boldsymbol{q}}_k$, ${\boldsymbol{\lambda}}_c$, ${\boldsymbol{\lambda}}_f$, and ${\boldsymbol{\lambda}}_a$. Notably, we directly define the normal contact force and friction force on the manifold of ${\boldsymbol{\lambda}}_n$ and ${\boldsymbol{\lambda}}_t$. An essential advantage of our method lies in the provision of a smooth system of nonlinear algebraic equations. Consequently, these equations can be solved using widely used techniques such as the gradient method or the Newton-Raphson method.
\section{Numerical simulation}\label{sec::simu}
\subsection{Influence of friction}
\subsubsection{Internal contact}
\begin{figure}[h]
	\centering
	\includegraphics[width=0.45\textwidth]{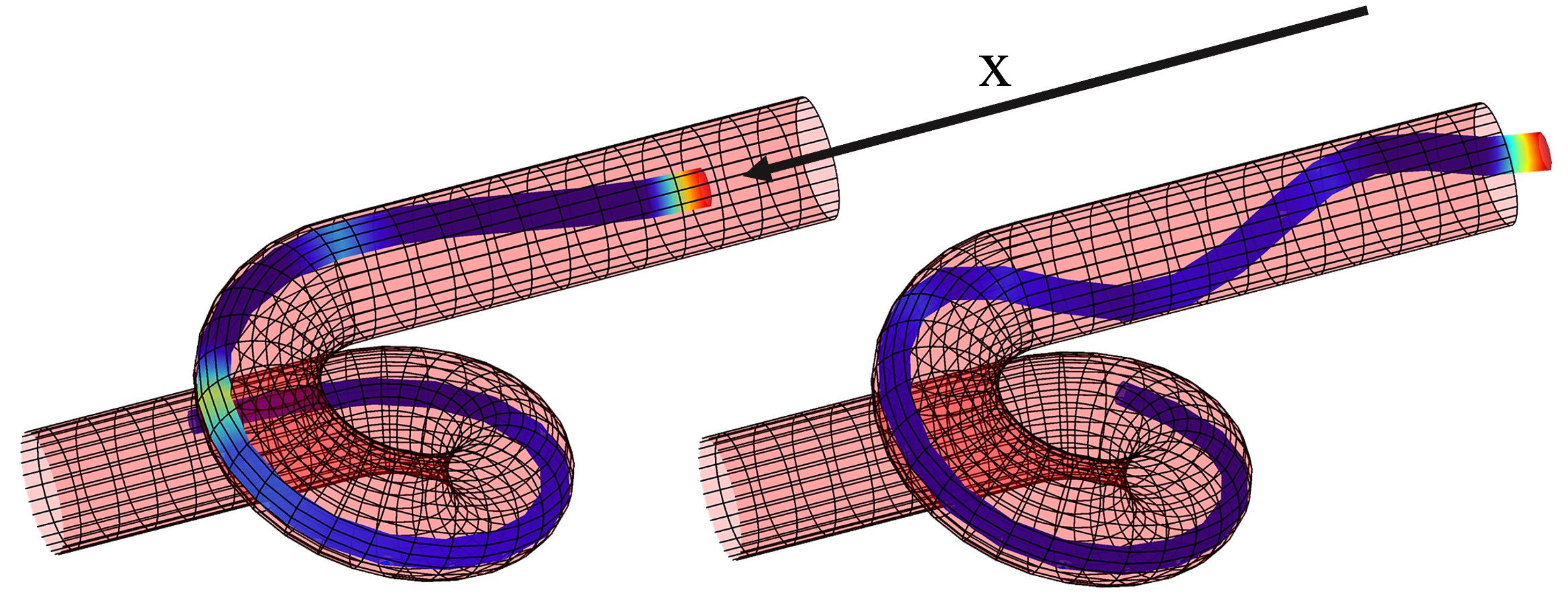}
	\caption{Fictional contact in a curved tube. Successful case(left) and buckling case(right).}
	\label{fig:tube}
\end{figure}

\begin{figure}[h]
	\centering
	\includegraphics[width=0.5\textwidth]{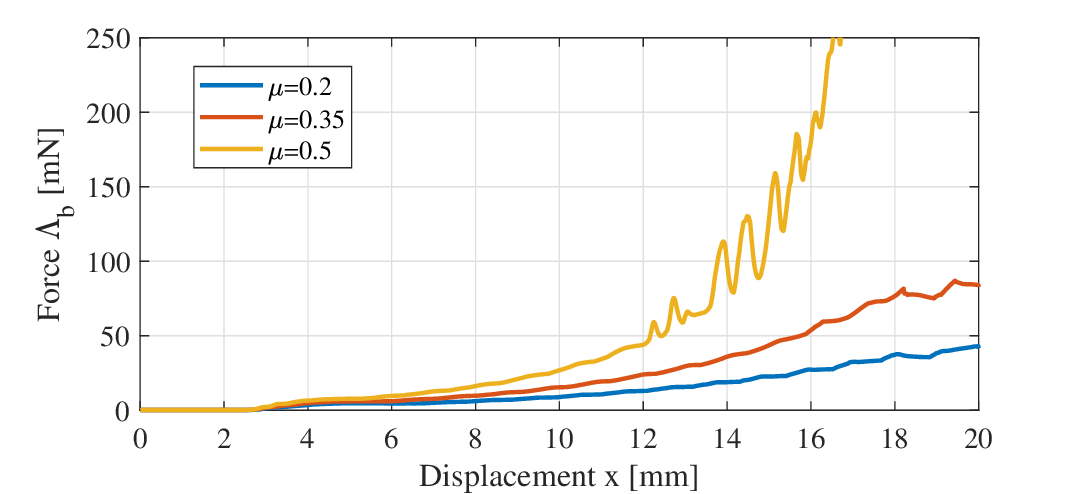}
	\caption{Fictional contact in a 3D  curved tube.}
	\label{fig:tubeforce}
\end{figure}

In this test, we insert a soft rod (blue) of length $\mathrm{L}=26\mathrm{mm}$ inside a rigid tube (red) along axis x. The diameter of soft rod is $0.3\mathrm{mm}$ at tip and $0.2\mathrm{mm}$ at end. The radius of curvature of the tube is $3\mathrm{mm}$. The Young modulus of soft rod is $55\mathrm{MPa}$ and Poisson ratio is $0.45$. Fig. \ref{fig:tube} shows the final state of insertion with different friction coefficients. The figure on the right side shows that the soft rod fails to be inserted and sticks inside the tube due to the large friction $\mu=0.5$, while the figure on left side shows the successful insertion with $\mu=0.2$. Fig. \ref{fig:tubeforce} shows the evolution of the norm of the constrained force $\boldsymbol{\Lambda}_{b}$ at tip with respect of the insertion displacement.
\begin{figure*}[h]
	\centering
	\includegraphics[width=1\textwidth]{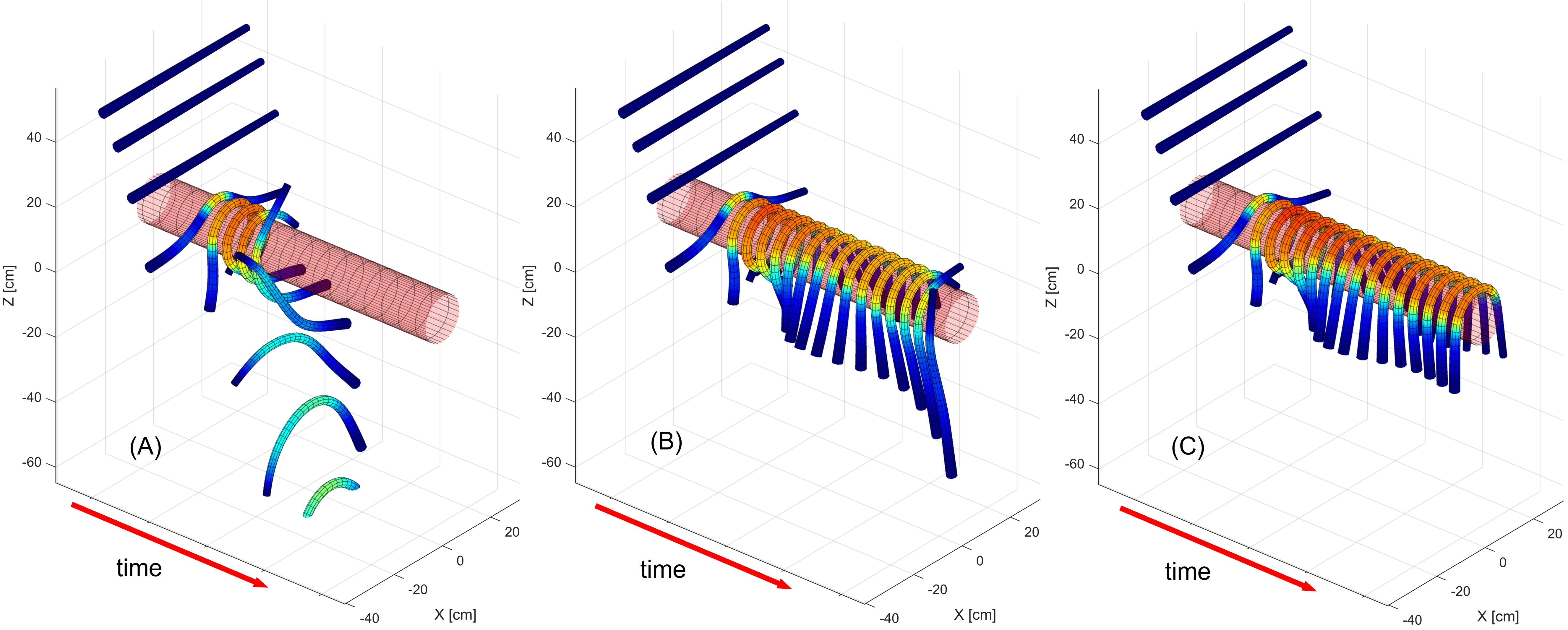}
	\caption{Fictional contact under gravity. The configuration of soft rod is plotted every $0.15s$. Number of sections: $n=m=20$. (A). friction coefficient $\mu=0$;  (B). friction coefficient $\mu=0.1$; (C). friction coefficient $\mu=0.2$. Implicit time step of simulation $\mathrm{dt}=0.005s$. The color shows the internal force of rods, which gradually increases from blue to red.}
	\label{fig:impact rod}
\end{figure*}
\subsubsection{External contact}
In this test, we throw a soft rod (blue) towards a rigid obstacle (red). The soft rod is initially positioned horizontally above the obstacle. The length of the soft rod is $30\mathrm{cm}$ and its diameter of two side of are $0.3\mathrm{cm}$ and $0.2\mathrm{cm}$ respectively. The Young modulus of soft rod is $55\mathrm{MPa}$ and Poisson ratio is $0.45$.  As shown in Fig. \ref{fig:impact rod}, we test the impact with different coefficient of friction. When the coefficient of friction is small, the soft rod cannot stay on the rigid rod after the collision occurs because there is not enough friction to resist the asymmetrical gravity. Meanwhile, if we increase the coefficient of friction, the soft robot will not slide down.
\subsection{Influence of discritization}
\begin{figure}[H]
	\centering
	\includegraphics[width=0.52\textwidth]{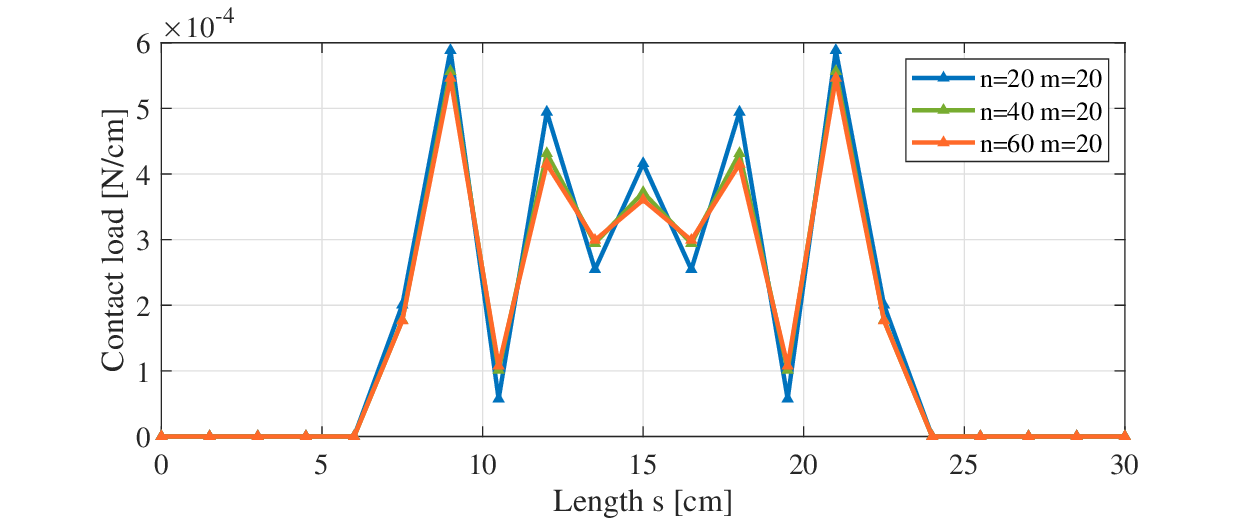}
	\caption{The influence of different number of section for strain interpolation on the distribution of contact load.}
	\label{fig:2bodyforce}
\end{figure}

In this subsection, we compare and analyze the impact of different discretization strategies on the computational results of our model. Specifically, we investigate the case of two intertwined rods initially positioned in a crossed configuration. The rods are then simultaneously twisted, resulting in mutual entanglement until a total twist angle of 540 degrees is reached.
\begin{figure}[H]
	\centering
	\includegraphics[width=0.45\textwidth]{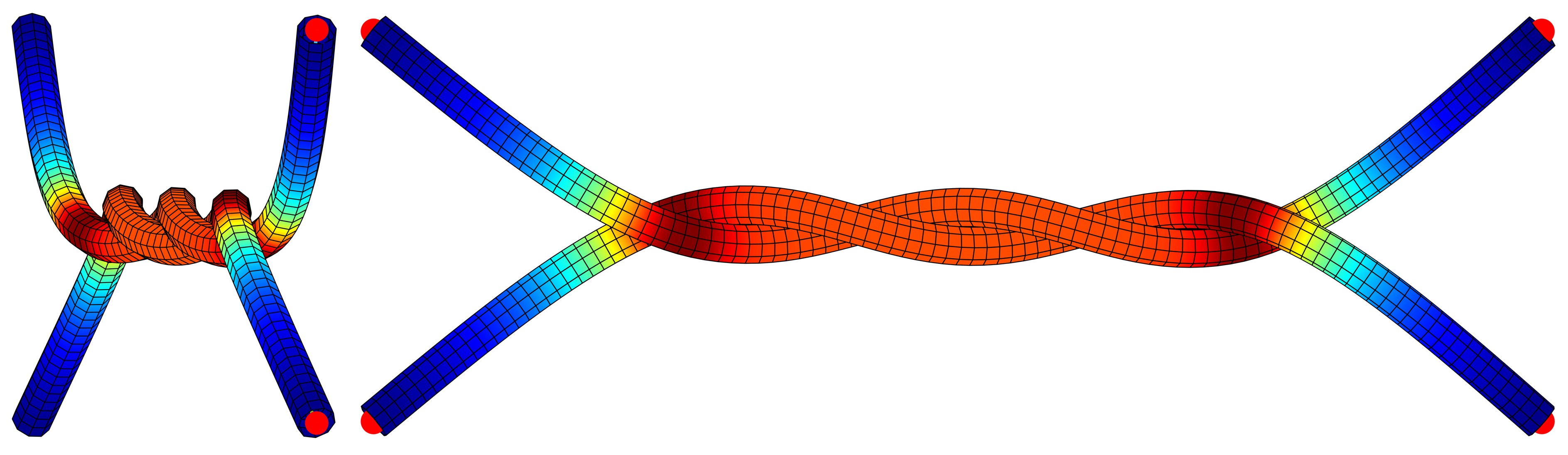}
	\caption{Two rods entangled with each other. The color shows the internal force of rods, which gradually increases from blue to red.}
	\label{fig:2body}
\end{figure}
We conduct several tests to examine the effects of different discretization strategies. Specifically, we fix the number of discrete sections ($m$) for the contact field and vary the number of discrete sections ($n$) for the strain field.
As shown in Fig. \ref{fig:2bodyforce}, we compare the results obtained from different discretization strategies on the contact load. This analysis allows us to understand the trade-offs between accuracy and computational efficiency in our model.
Through this comparative analysis, we aim to identify the optimal discretization strategy that balances computational efficiency with accurate representation of the physical phenomena involved in the entanglement process. The insights gained from this analysis will contribute to refining our model and enhancing its predictive capabilities. 

From Fig. \ref{fig:2bodyforce}, it can be observed that when the number of contact points remains constant, the discretization of the strain field affects the distribution of contact forces. With fewer sections in the strain field, the distribution of contact forces exhibits larger fluctuations. As the strain field is more finely discretized, the distribution of contact forces becomes smoother and the results tend to converge. Therefore, further discretization is meaningless at this point. 

Based on the above analysis, we can conclude that the ratio between the discretization of the strain field and the contact field is a key factor influencing computational efficiency. When the number of contact points remains constant, excessively fine discretization of the strain field does not significantly affect the computed contact forces. Additionally, the strain field should be discretized more finely than the contact field to ensure that the soft slender robot has enough degrees of freedom to satisfy the contact constraints. Based on simulation experience, we can conclude that a reasonable compromise is to have twice as many sections in the strain field compared to the contact field.

By considering these findings, we can optimize the discretization strategy for our model, achieving a balance between computational efficiency and accuracy in predicting the contact forces and capturing the overall behavior of the soft robot system.
\subsection{Choice of smooth function}
\begin{figure}[h]
	\centering
	\includegraphics[width=0.5\textwidth]{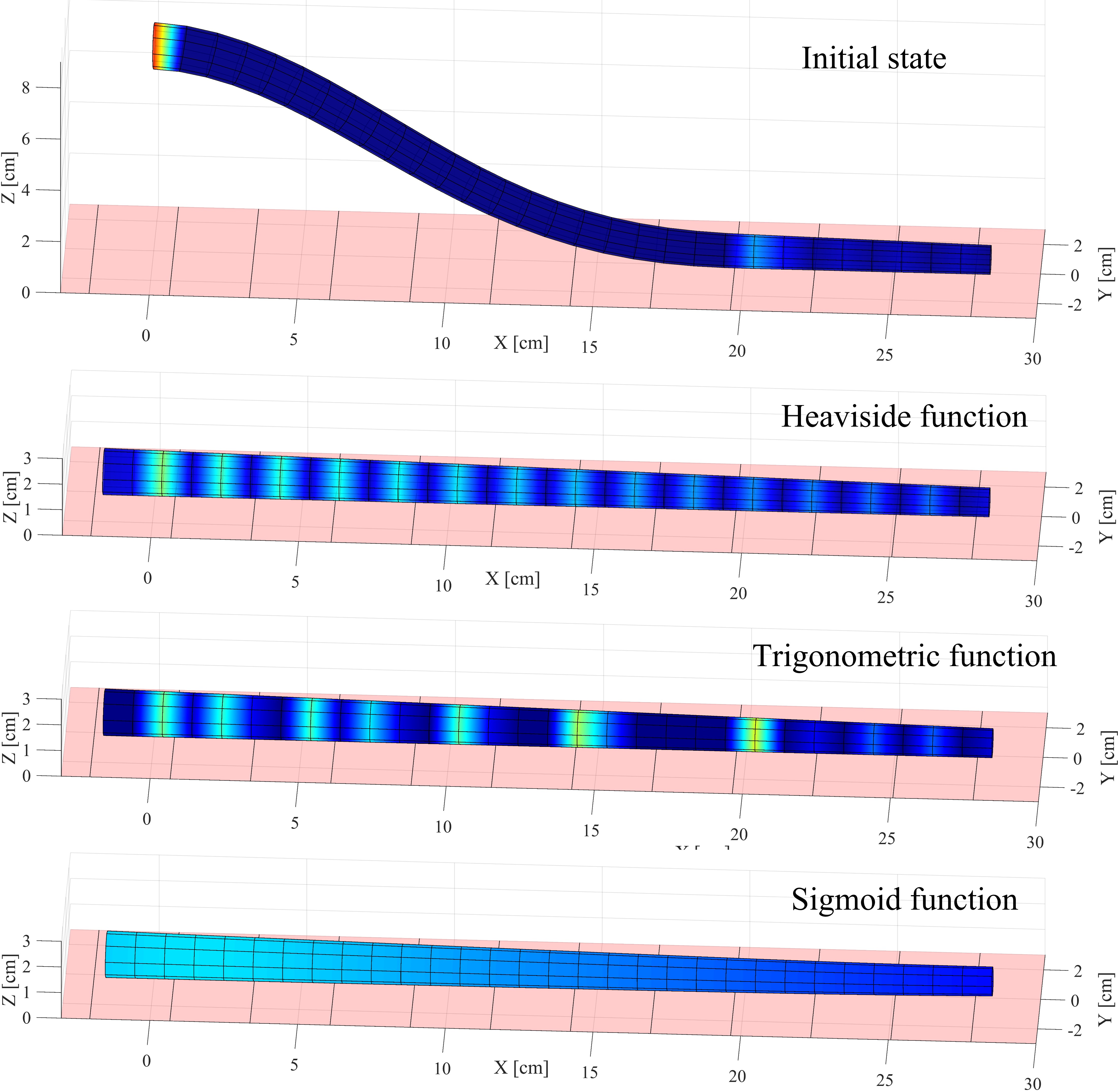}
	\caption{Final steady state calculated using different functions. The colors in the plot represent the contact load distribution.}
	\label{fig:choicecontact}
\end{figure}

\begin{figure}[h]
	\centering
	\includegraphics[width=0.52\textwidth]{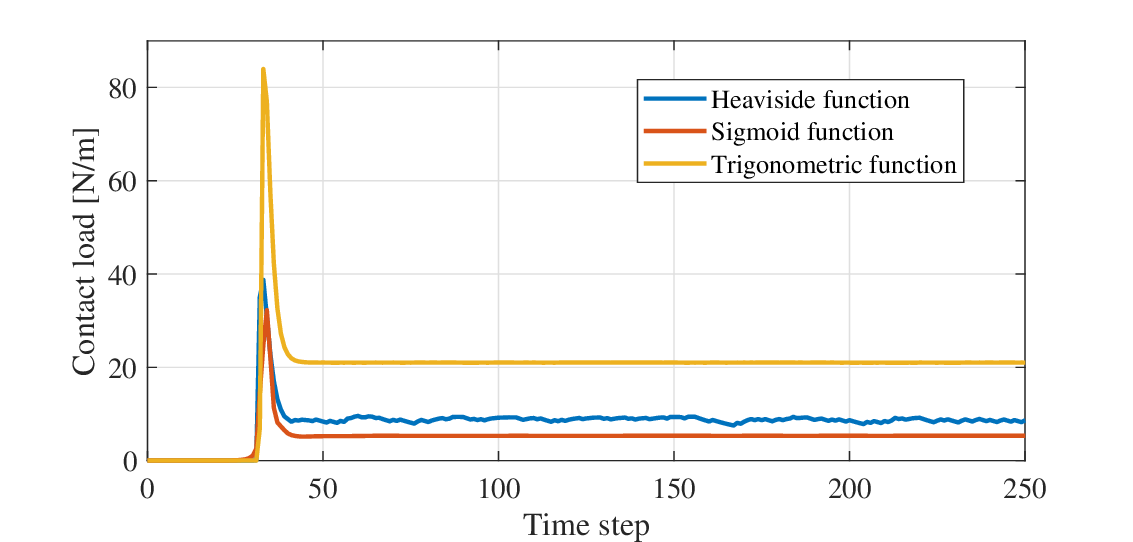}
	\caption{Evolution of contact loads over time step iterations. The point of contact is 6 cm from the left end. The time step is set as $0.01s$.}
	\label{fig:choice}
\end{figure}

In Section \ref{smoothfunc}, we mentioned that the Heaviside function used to construct the complementarity conditions can be approximated or equivalently represented by other continuous or smooth functions. Through simulations, we will explore the influence of different approximation functions on the solution results. Fig. \ref{fig:choicecontact} illustrates the case of a homogeneous tapered slender robot placed on a plane. In the initial state, the left end of the slender robot is suspended and fixed by a constraint force, while the right end contacts the ground under the influence of gravity. The slender robot has a length of $30 \mathrm{cm}$, density of $3\times10^3 \mathrm{kg}/\mathrm{m}^3$, Young's modulus of $1\mathrm{Mpa}$ and friction coefficient of $0.5$. After the initial state, the left end of the slender robot is released and starts to fall until it collides with the ground, reaching a final equilibrium state. Fig.\ref{fig:choicecontact} shows the steady states obtained using the impact function and different smoothing functions. In all three test groups, the strain field of the slender robot is divided into 30 sections, as well as the contact field.
From the distribution of contact loads shown in Fig. \ref{fig:choicecontact}, it can be observed that for the Heaviside function and trigonometric function, which satisfy the complementarity constraints exactly, although the slender robot is in a force-balanced state and remains stationary, not all contact points are activated but some contact points are in a critical state (virtual contact). In this case, due to the discretization of the system, the solution of the mechanical equation system is not unique. For the Heaviside function, as shown in fig.\ref{fig:choice}, the contact force undergoes small jumps during the iteration process, which is also due to the presence of virtual contacts. Since the slack variables associated with virtual contacts are zero, the Heaviside function is discontinuous, resulting in numerical jumps during the iteration. However, this does not occur for the trigonometric function, which remains smooth and continuous. Among the three simulations, the sigmoid function yields results closest to the real solution. As shown in Fig. \ref{fig:choicecontact}, the contact load is smoothly distributed and remains stable during the iteration process. {This is because the sigmoid function is an approximation of the complementarity constraint. In this case, there are no virtual contacts where both the contact gap and contact force are zero, and therefore, all contact points are activated, yielding a unique solution.}

Based on these findings, we conclude that approximating the impact function with trigonometric functions can address the issue of numerical jumps in critical states. However, in some cases, non-uniqueness of solutions may arise. On the other hand, using the sigmoid function, although it cannot precisely satisfy the complementarity constraint in the vicinity of critical states, offers better robustness.
\section{Experiment for model validation}\label{sec::exp}
\subsection{Experiment setup}
\begin{figure}[h]
	\centering
	\includegraphics[width=0.35\textwidth]{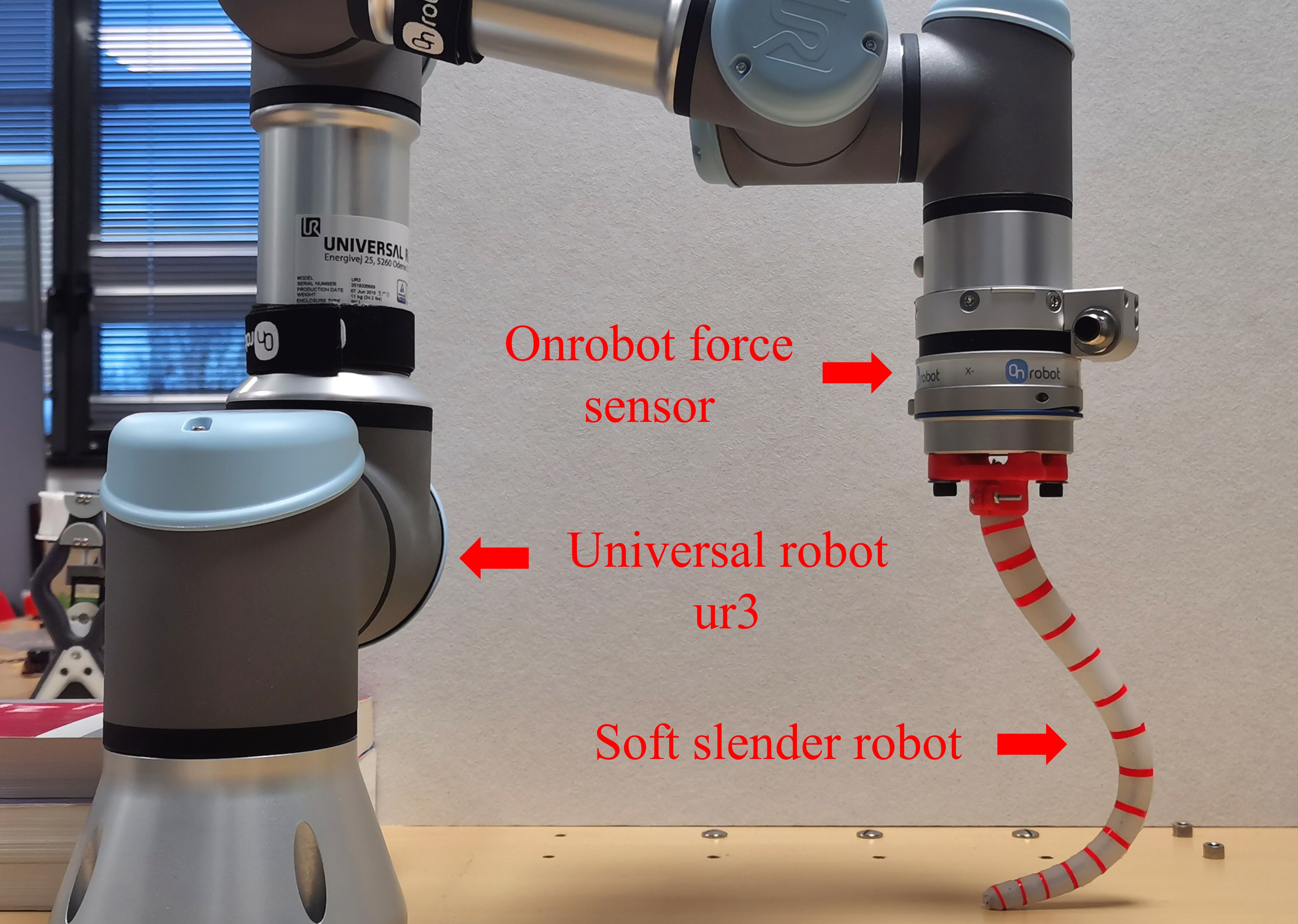}
	\caption{Experiment setup}
	\label{fig:setup}
\end{figure}

As shown in fig.\ref{fig:setup}, our platform contains the soft silicone robot whose base is fixed with the rigid manipulator (Universal robot ur3). There is a steel string passing through the center line of the soft silicone robot in order to increase its stiffness. We installed force sensor (Onrobot) at the end of the rigid manipulator to measure the contact force of the soft silicone robot. Specific image processing program has been developed by us to track the red plastic markers on the soft silicone robot to interpolate the shape of polymer actuator and compute the corresponding curvatures at each control points, which will be detailed in the next subsection. TABLE \ref{tab:Measured parameters} shows the measured physic parameters of the soft silicone robot and steel string. In general, soft silicone robots can be equipped with various actuators, such as cable-driven, tendon-driven, and magnetic-driven mechanisms. Our study introduces a contact dynamics framework that is suitable for analyzing soft slender robots with these actuators. {However, in our research, we intentionally focused on validating the deformation and contact force magnitude of the soft slender robot during interactions with the external environment. Consequently, we designed the robot under investigation without any actuation structure and only passive mode (i.e. no actuation) is used to validate the proposed modeling framework. The investigation of contact issues in actuated soft slender robots will be addressed in our future work.}
\begin{table}[h]
	\centering
	\caption{Physic parameters of silicone robot and steel string}
	\label{tab:Measured parameters}
	\setlength{\tabcolsep}{1mm}{
		\begin{tabular}{|c|c|c|}
			\hline
			& & \\[-6pt]
			Length $L$&Silicone robot&Steel string\\
			\hline
			& & \\[-6pt]
			Radius $R_0$&0.85 cm &0.15 cm\\
			\hline
			& & \\[-6pt]
			Radius $R_1$&0.5 cm&0.15 cm\\
			\hline
			& & \\[-6pt]
			Poisson ratio $\gamma$ &0.45&0.3\\
			\hline
			& & \\[-6pt]
			Young's modulus $E$ &2.56 $\times 10^{5}$Pa&1.2 $\times 10^{9}$Pa\\
			\hline
			& & \\[-6pt]
			Mass density $\rho$ &1.41$\times$10$^{3}$ kg/m$^3$&7.8$\times$10$^{3}$ kg/m$^3$\\
			\hline
	\end{tabular}}
\end{table}
\subsection{Strain observer design}
To compare the strain of a real robot with our proposed model, we initially develop a strain observer that estimates the strain of the real robot based on the positions of markers.

By analyzing the images captured by a rapid camera, we can extract the shape of the actuator, enabling us to compute the strain along the soft robot. We employ a combination of image binarization, inter-frame difference \cite{cheng2014motion}, and center point extraction algorithms in Matlab to obtain the marker positions. 

We can conceptualize the presence of a virtual soft robot with zero inertia that consistently mimics the deformation of the real soft robot. This can be viewed as the virtual soft robot passing through the real robot while being constrained by the measured positions of the markers using the bilateral constraint discussed in Section \ref{sec::Bilateral_control_constrain}. In this context, the real soft robot and the virtual robot correspond to two contact objects, while the markers serve as the fixed contact constraints. Thus the design of observer can be formulated as a contact problem. {Specifically, according to the principle of minimum potential energy, the total potential energy of the virtual robot should hold the minimum under the contact constraints of markers, which leads to a minimization problem with constraints:
\begin{equation}\label{minobs}
	\operatorname*{argmin}_{\hat{\boldsymbol{q}}}\ \frac{1}{2}\hat{\boldsymbol{q}}^\top\boldsymbol{K}\hat{\boldsymbol{q}}+V_g(\hat{\boldsymbol{q}})
\end{equation}
\begin{equation}\label{key}
	\mathrm{subject \ to} \ \boldsymbol{G}_{f}(\hat{\boldsymbol{q}})=\mathbf{0}
\end{equation}
where $\hat{\boldsymbol{q}}$ is the estimated states, representing the strain field of the virtual robot. $\boldsymbol{K}$ stands for the positive diagonal stiffness matrix and $V_g(\hat{\boldsymbol{q}})$ stands for the gravitational potential energy of the virtual robot. $\boldsymbol{G}_{f}(\hat{\boldsymbol{q}})$ contents the constraints which are defined in (\ref{gf}) by setting the configurations of markers as the fixed contact constraints.
Then, we can find the solution of this minimization problem by solving its Karush-Kuhn-Tucker Conditions:  
\begin{subnumcases}{\label{governing3}}
	\boldsymbol{K}\hat{\boldsymbol{q}}+\boldsymbol{P}(\hat{\boldsymbol{q}})-\boldsymbol{H}_{f}(\hat{\boldsymbol{q}})\boldsymbol{\Lambda}_{f}=\mathbf{0}\label{dyn3}
	\\
	\boldsymbol{G}_{f}(\hat{\boldsymbol{q}})=\mathbf{0}
\end{subnumcases}
where $\boldsymbol{P}(\hat{\boldsymbol{q}})$ is defined in (\ref{dynamic}), meaning the gradient of $V_g(\hat{\boldsymbol{q}})$. $\boldsymbol{H}_{f}(\hat{\boldsymbol{q}})$ is defined in (\ref{dynamic}), meaning the transpose of the Jacobian of $\boldsymbol{G}_{f}(\hat{\boldsymbol{q}})$. $\boldsymbol{\Lambda}_{f}$ is the Lagrange multiplier, meaning the contact force. }
\subsection{Sticking test}
\begin{figure}[h]
	\centering
	\includegraphics[width=0.42\textwidth]{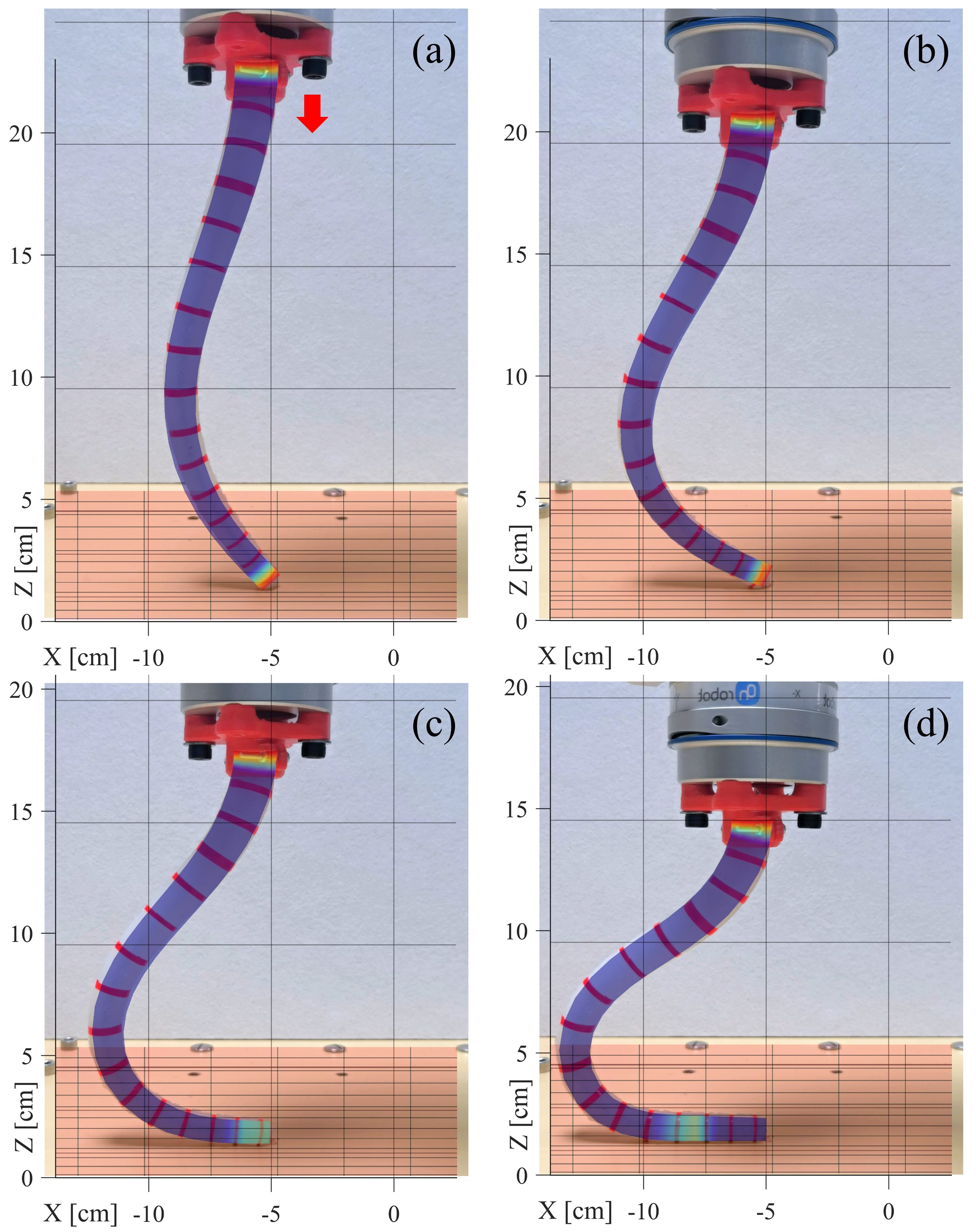}
	\caption{Configurations of sticking test and simulation.}
	\label{fig:sticktest}
\end{figure}

\begin{figure}[h]
	\centering
	\includegraphics[width=0.53\textwidth]{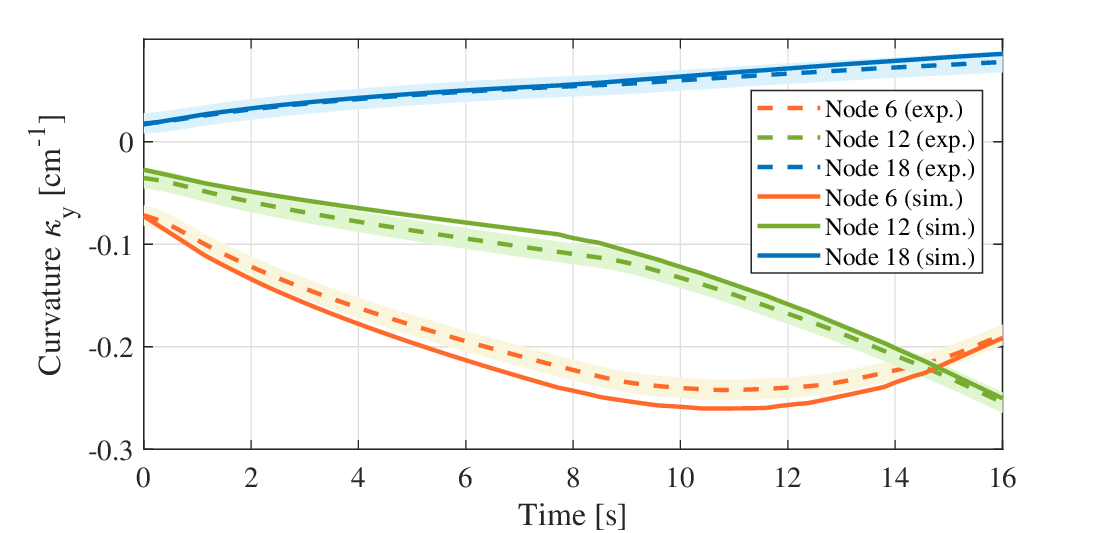}
	\caption{Evolution of strain of sticking test and simulation. Shaded area represents measurement errors.}
	\label{fig:stickstrain}
\end{figure}

\begin{figure}[h]
	\centering
	\includegraphics[width=0.5\textwidth]{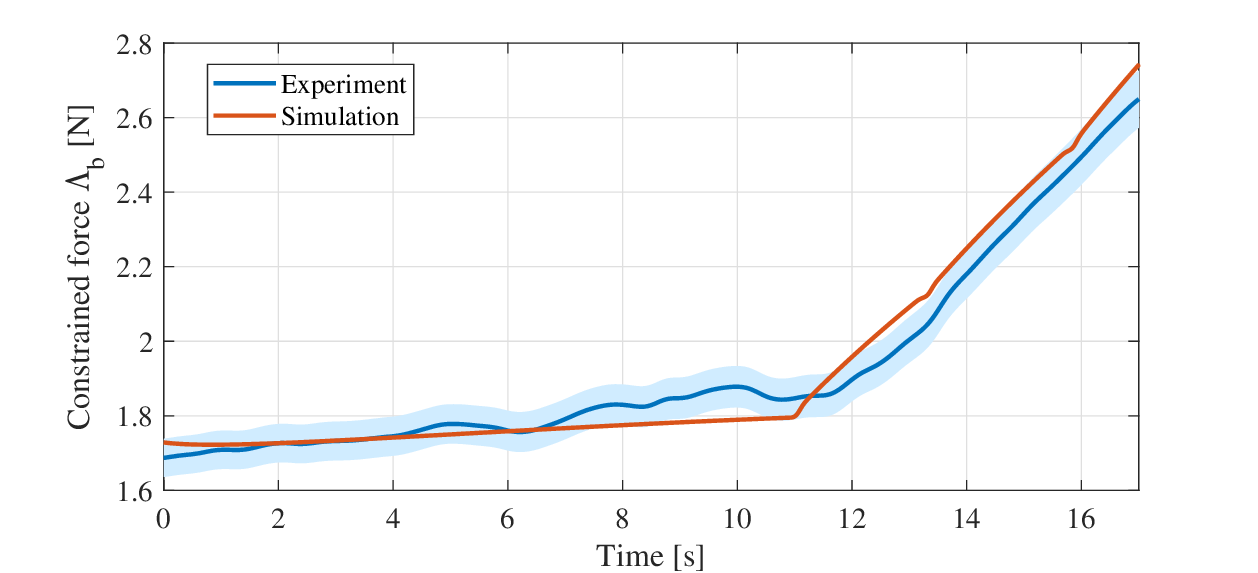}
	\caption{Evolution of contact force of sticking test and simulation. Shaded area represents measurement errors.}
	\label{fig:stickforce}
\end{figure}

We first valid our proposed model through comparing the deformation and contact force both obtained from experiment and simulation under the stick state. As shown in Fig. \ref{fig:sticktest}, the base of the soft robot is pressed slowly by the rigid manipulator while the tip of the soft robot is always at stick contact state with the planar of table due to the friction during the total process. We compare the deformation of experiment and simulation through the observed strain of real soft robot and the strain of simulation. {To determine the coefficient of friction between the silicone and the tabletop surface, we conducted experiments using silicone block made from the same material as the robot. The block was placed on the tabletop, and a pushing force was applied to initiate movement. By measuring the ratio of the applied force to the weight of the silicone gel blocks, we were able to calculate the coefficient of friction. After performing several experiments, we obtained a measured coefficient of friction of $0.83$ between the silicone block and the tabletop surface, which will be used in the simulation.} Fig. \ref{fig:stickstrain} shows the evolution of strain at control points. Fig. \ref{fig:stickforce} shows the evolution of the total normal contact force of soft robot. \textcolor{black}{The test is repeated 5 times and the average error percentages of both strain and force from tests are less than $10 \% $.}
\subsection{Sliding test}
\begin{figure}[h]
	\centering
	\includegraphics[width=0.43\textwidth]{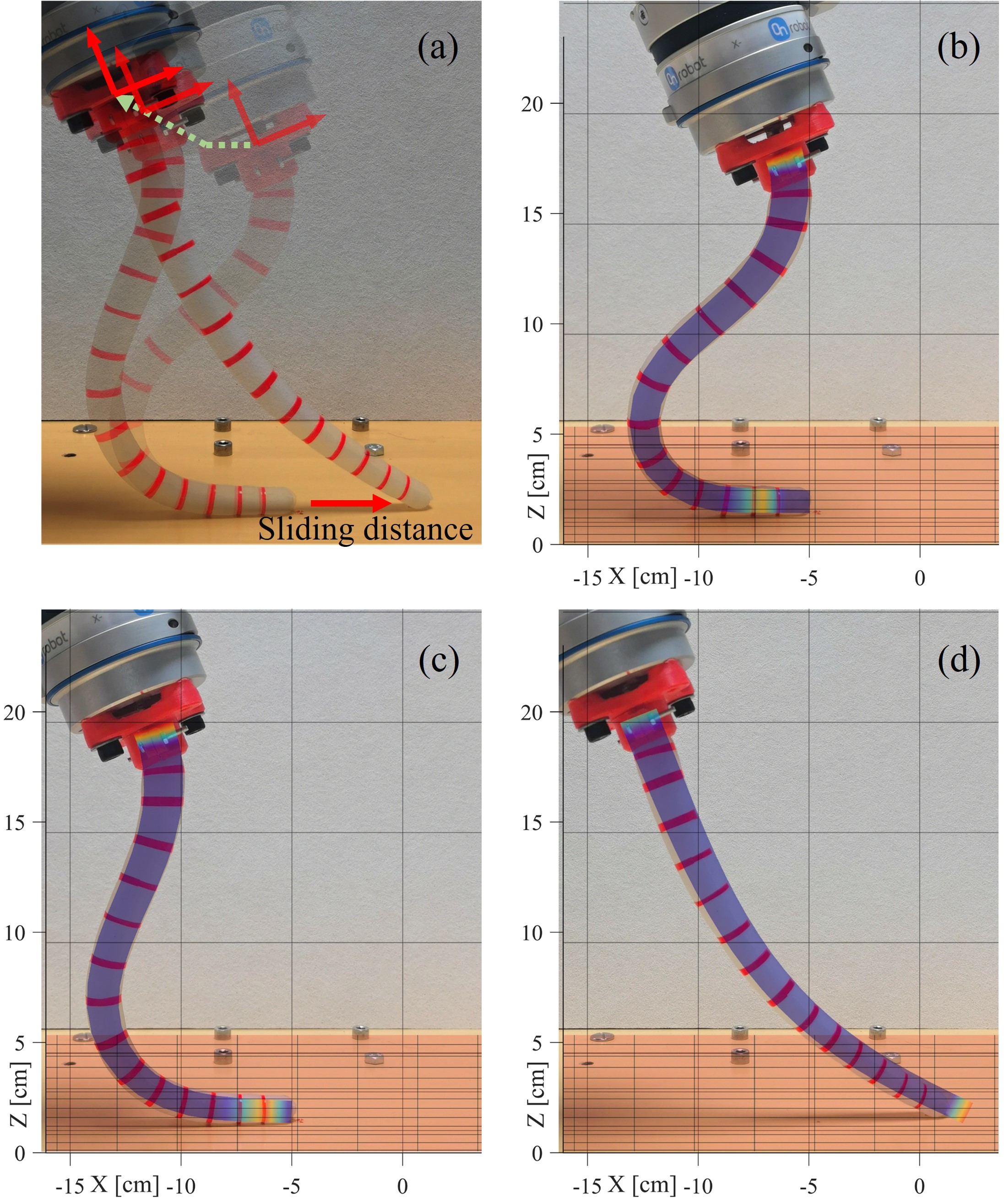}
	\caption{Configurations of sliding test and simulation.}
	\label{fig:sliding}
\end{figure}

\begin{figure}[h]
	\centering
	\includegraphics[width=0.5\textwidth]{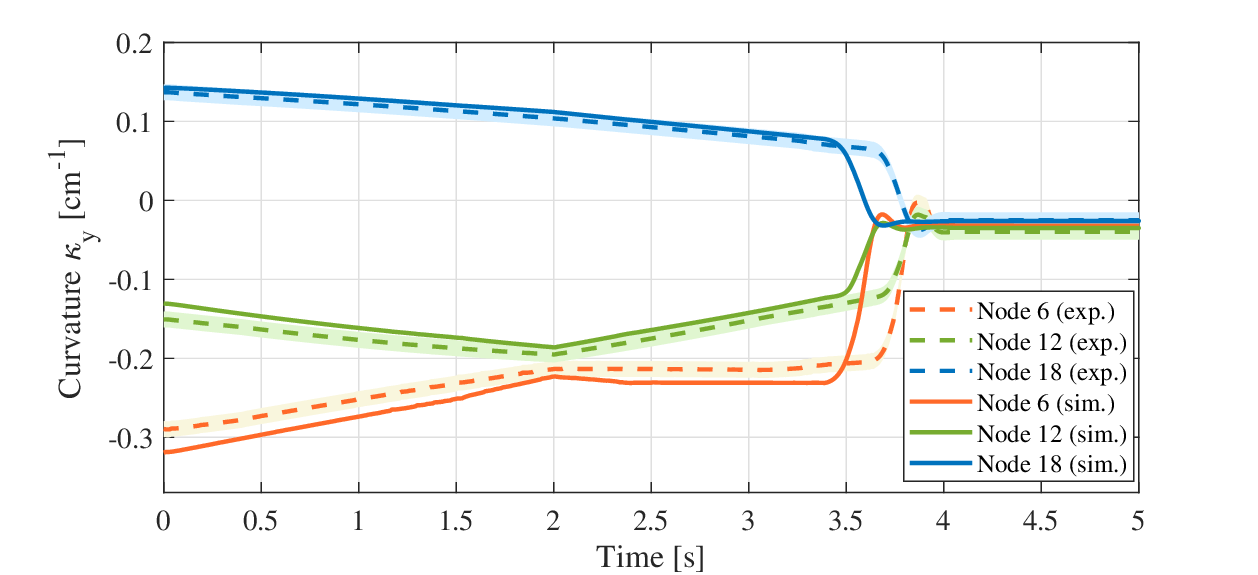}
	\caption{Evolution of strain of sliding test and simulation. Shaded area represents measurement errors.}
	\label{fig:slidestrain}
\end{figure}

\begin{figure}[t]
	\centering
	\includegraphics[width=0.5\textwidth]{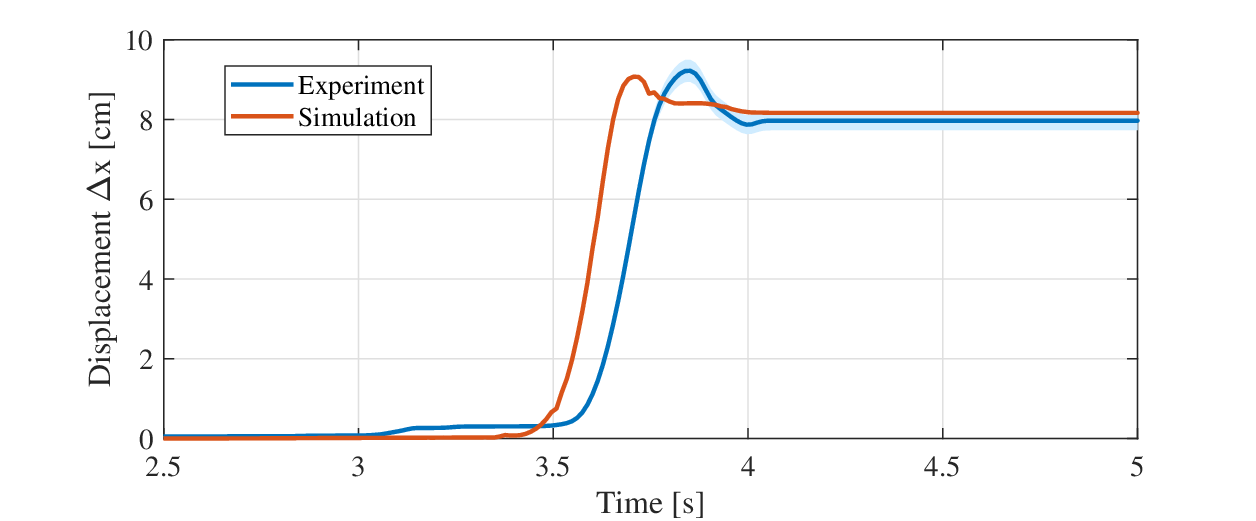}
	\caption{Evolution of sliding displacement of sliding test and simulation. Shaded area represents measurement errors.}
	\label{fig:slidex}
\end{figure}

Our second test aims to valid the proposed model in aspect of dynamic evolution from sticking state to sliding state. In this test, for the sake of a obvious conversion phenomena, we decrease the coefficient of friction by lubricating some lube on the surface of table. The coefficient of friction is measured as 0.48. As shown in Fig. \ref{fig:sliding}(a), the soft robot is first pre-curved by the rigid manipulator and is at stick state.  Then the rigid manipulator moves the base of soft robot away from the sticking contact point following the planned trajectory. The change of configuration of soft robot will lead to the transition of contact between soft robot and table from stick to slide.
Fig. \ref{fig:sliding}(b)-(d) shows the dynamic evolution of this phenomena during the experiment on comparing with the simulation result. Fig. \ref{fig:slidestrain} shows the evolution of strain at control points. Fig. \ref{fig:slidex} shows the evolution of the slide distance of tip. \textcolor{black}{The test is repeated 5 times and the average error percentages of both strain and sliding distance from texts are less than $15 \% $.}
\section{Conclusion}\label{sec::conc}
In this article, we propose a contact dynamics model for soft robots based on the Cosserat theory. The contact dynamics formulation considers the geometric relationships, forces, and moments at the contact points, allowing for the analysis of contact forces, frictional forces, and their effects on the robot's dynamics.

The contact constraints are formulated as a set of equations that ensure non-penetration, force equilibrium, and frictional interactions at the contact interfaces. We reformulate the contact constraints by transforming them into equality constraints and smoothing the non-smooth constraints. The reconstructed contact dynamics system can be solved directly using general-purpose numerical methods such as the Newton's method, and it exhibits good robustness.

By incorporating the Cosserat theory into the contact dynamics model, we achieve a more comprehensive and accurate representation of the contact interactions in soft robots. This enables us to study and analyze the behavior and performance of soft robots under different contact conditions, facilitating the design and control of soft slender robot systems for various applications.  {In our future endeavors, we will expand upon the contact dynamics framework we have proposed and shift our focus towards modeling and analyzing soft slender robots that incorporate actuators, such as cable-driven, tendon-driven, and magnetic-driven mechanisms. Specifically, we will tackle the control challenges that arise when soft slender robots interact with the external environment. By incorporating these drive systems, our aim is to enhance the understanding and capabilities of soft slender robot control in a wide range of real-world applications.}

Overall, the proposed contact dynamics model based on the Cosserat theory provides a valuable framework for understanding and simulating the contact behavior of soft robots, advancing the field of soft slender robotics and its applications.
\nocite{*}  
\bibliographystyle{IEEEtran}  
\bibliography{contact}  

\end{document}